\def\year{2021}\relax
\documentclass[letterpaper]{article} 
\usepackage{aaai21}  
\usepackage{times}  
\usepackage{helvet} 
\usepackage{courier}  
\usepackage[hyphens]{url}  
\usepackage{graphicx} 
\usepackage{natbib}  
\usepackage{caption} 
\usepackage{comment}
\usepackage{amsmath}
\usepackage{booktabs}
\usepackage{graphicx}
\usepackage{multicol}
\usepackage{multirow}
\usepackage{textcomp}
\usepackage{pifont}

\graphicspath{ {./figures/} }

\newcommand{\code}[1]{\texttt{#1}}

\newcommand{\costequivalentcurve}[0]{cost equivalent curve}
\newcommand{\Costequivalentcurve}[0]{Cost equivalent curve}

\newcommand{\dataset}[1]{\textsc{#1}}

\newcommand{\anli}[0]{\dataset{$\alpha$NLI}}
\newcommand{\cosmosqa}[0]{\dataset{CosmosQA}}
\newcommand{\hellaswag}[0]{\dataset{HellaSWAG}}
\newcommand{\piqa}[0]{\dataset{PIQA}}
\newcommand{\physicaliqa}[0]{\piqa{}}
\newcommand{\socialiqa}[0]{\dataset{SocialIQa}}
\newcommand{\winogrande}[0]{\dataset{WinoGrande}}
\newcommand{\commonsenseqa}[0]{\dataset{CommonsenseQA}}
\newcommand{\joci}[0]{\dataset{JOCI}}
\newcommand{\cycic}[0]{\dataset{CycIC}}

\newcommand{\glue}[0]{\dataset{GLUE}}
\newcommand{\superglue}[0]{\dataset{SuperGLUE}}
\newcommand{\rainbow}[0]{\dataset{Rainbow}}

\newcommand{\atomic}[0]{\dataset{Atomic}}
\newcommand{\conceptnet}[0]{\dataset{ConceptNet}}

\newcommand{\none}[0]{\dataset{None}}
\newcommand{\together}[0]{\dataset{Both}}

\newcommand{\model}[1]{\mbox{\textsc{#1}}}

\newcommand{\tfive}[1]{\model{T5{#1}}}
\newcommand{\universalmodel}[1]{\model{Unicorn{#1}}}

\makeatletter
\let\ftype@table\ftype@figure
\makeatother

\title{\universalmodel{} on \rainbow{}: \\
A Universal Commonsense Reasoning Model on a New Multitask Benchmark}
\author{
Nicholas Lourie\textsuperscript{\rm $\spadesuit$} \,
Ronan Le Bras\textsuperscript{\rm $\spadesuit$} \,
Chandra Bhagavatula\textsuperscript{\rm $\spadesuit$} \,
Yejin Choi \textsuperscript{\rm $\heartsuit$}\textsuperscript{\rm $\spadesuit$}
}
\affiliations {
    \\
    \textsuperscript{\rm $\spadesuit$} Allen Institute for AI, WA, USA\\
    \textsuperscript{\rm $\heartsuit$} Paul G. Allen School of Computer Science \& Engineering, WA, USA \\
}

\begin{document}

\maketitle

\begin{abstract}
Commonsense AI has long been seen as a near impossible goal---until recently. Now, research interest has sharply increased with an influx of new benchmarks and models.

We propose two new ways to evaluate commonsense models, emphasizing their generality on new tasks and building on diverse, recently introduced benchmarks. First, we propose a new multitask benchmark, \rainbow{}, to promote research on commonsense models that generalize well over multiple tasks and datasets. Second, we propose a novel evaluation, the \textbf{\costequivalentcurve{}}, that sheds new insight on how the choice of source datasets, pretrained language models, and transfer learning methods impacts performance and \emph{data efficiency}.

We perform extensive experiments---over 200 experiments encompassing 4800 models---and report multiple valuable and sometimes surprising findings, e.g., that transfer almost always leads to better or equivalent performance if following a particular recipe, that QA-based commonsense datasets transfer well with each other, while commonsense knowledge graphs do not, and that perhaps counter-intuitively, larger models benefit more from transfer than smaller ones.

Last but not least, we introduce a new universal commonsense reasoning model, \universalmodel{}, that establishes new state-of-the-art performance across 8 popular commonsense benchmarks, \anli{} ($\to$\textbf{87.3\%}), \cosmosqa{} ($\to$\textbf{91.8\%}), \hellaswag{} ($\to$\textbf{93.9\%}), \physicaliqa{} ($\to$\textbf{90.1\%}), \socialiqa{} ($\to$\textbf{83.2\%}), \winogrande{} ($\to$\textbf{86.6\%}), \cycic{} ($\to$\textbf{94.0\%}) and \commonsenseqa{} ($\to$\textbf{79.3\%}).
\end{abstract}

\section{Introduction}
\label{sec:introduction}

\begin{figure}[h]
    \centering
    \includegraphics[width=\columnwidth]{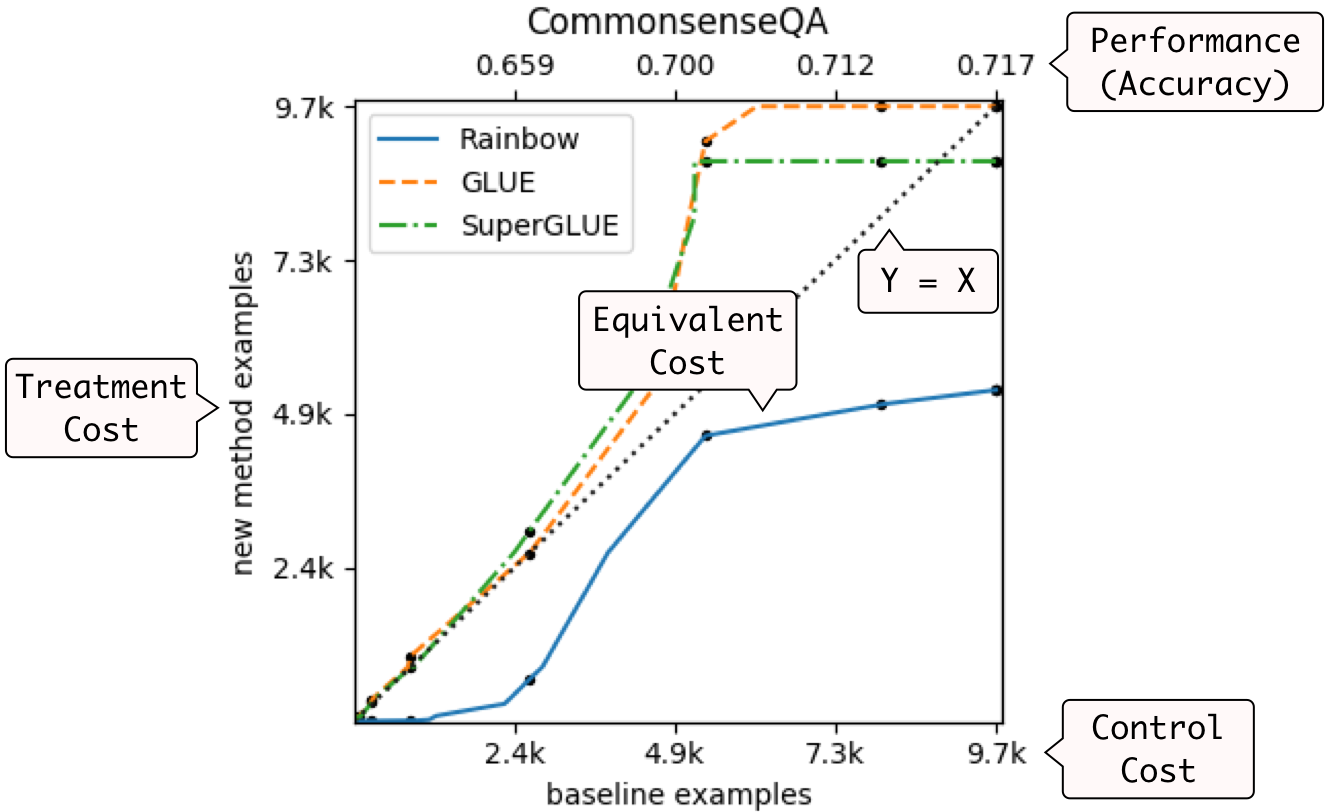}
    \caption{\textit{\Costequivalentcurve{}s} comparing transfer learning from \glue{}, \superglue{}, and \rainbow{} onto \commonsenseqa{}. Each curve plots how much  training data the single-task baseline (the $x$-axis) needs compared to the multitask method (the $y$-axis) to achieve the same performance (shown on the top axis in accuracy). \emph{Curves below the diagonal line ($y=x$) indicate that the multitask method needs less training data from the target dataset than the single-task baseline for the same performance.} Thus, lower curves mean more successful transfer learning.}
    \label{fig:equivalent-curve}
\end{figure}

In AI's early years, researchers sought to build machines with common sense \citep{mccarthy59a}; however, in the following decades, common sense came to be viewed as a near impossible goal. It is only recently that we see a sudden increase in research interest toward commonsense AI, with an influx of new benchmarks and models \citep{mostafazadeh-etal-2016-corpus, talmor-etal-2019-commonsenseqa, Sakaguchi2020WINOGRANDEAA}. 

This renewed interest in common sense is ironically encouraged by both the great empirical strengths and limitations of large-scale pretrained neural language models. On one hand, pretrained models have led to remarkable progress across the board, often surpassing human performance on leaderboards \citep{radford2018improving, devlin-etal-2019-bert, liu2019roberta, 2019t5}. On the other hand, pretrained language models continue to make surprisingly silly and \emph{nonsensical} mistakes, even the recently introduced GPT-3.\footnote{\url{https://www.technologyreview.com/2020/08/22/1007539/gpt3-openai-language-generator-artificial-intelligence-ai-opinion/}} This motivates new, relatively under-explored research avenues in commonsense knowledge and reasoning.

In pursuing commonsense AI, we can learn a great deal from mainstream NLP research. In particular, the introduction of multitask benchmarks such as \glue{} \cite{wang2019glue} and \superglue{} \cite{wang2019superglue} has encouraged fundamental advances in the NLP community, accelerating research into models that robustly solve many tasks and datasets instead of overfitting to one in particular. In contrast, commonsense benchmarks and models are relatively nascent, thus there has been no organized effort, to date, at administering a collection of diverse commonsense benchmarks and investigating transfer learning across them.

We address exactly this need, proposing two new ways to evaluate commonsense models with a distinct emphasis on their generality across tasks and domains. First, we propose a new multi-task benchmark, \rainbow{}, to facilitate research into commonsense models that generalize well over multiple different tasks and datasets. Second, we propose a novel evaluation, the \textbf{\costequivalentcurve{}}, that sheds new insight on how different choices of source datasets, pretrained language models, and transfer learning methods affect performance and data efficiency in the target dataset.

The primary motivation for \costequivalentcurve{}s is \textbf{data efficiency}. The necessary condition for state-of-the-art neural models to maintain top performance on any dataset is a sufficiently large amount of training data for fine-tuning. Importantly, building a dataset for a new task or a domain is an expensive feat, easily costing tens of thousands of dollars \citep{zellers2018swagaf}. Therefore, we want the models to \emph{generalize systematically} across multiple datasets, instead of relying solely on the target dataset.

Shown in Figure~\ref{fig:equivalent-curve}, the \costequivalentcurve{} aims to answer the following intuitive question: \emph{how much data does a transfer learning approach save over the baseline that doesn't benefit from transfer learning?} We provide a more detailed walk-through of this chart in \S\ref{sec:equivalent-curves}. As will be seen, \costequivalentcurve{}s have distinct advantages over simple evaluations at the full dataset size or classical learning curves drawn for each method and dataset separately, as they provide more accurate comparative insights into data efficiency in the context of multitasking and transfer learning.

We leverage these new tools to reevaluate common approaches for \textit{intermediate-task transfer} \citep{pruksachatkun2020intermediate}. Through extensive experiments, we identify multiple valuable and sometimes surprising findings, e.g., that intermediate-task transfer can always lead to better or equivalent performance if following a particular recipe, that QA-based commonsense datasets transfer well to each other, while commonsense knowledge graphs do not, and that perhaps counter-intuitively, larger models benefit much more from transfer learning compared to smaller ones.

In addition to the empirical insights, we also introduce a new universal commonsense reasoning model: \universalmodel{}, establishing new state-of-the-art performances across 8 benchmarks: \anli{} (\textbf{87.3\%}) \citep{bhagavatula2020abductive}, \cosmosqa{} (\textbf{91.8\%}) \citep{Huang2019CosmosQM}, \hellaswag{} (\textbf{93.9\%}) \citep{Zellers2019HellaSwagCA}, \physicaliqa{} (\textbf{90.1\%}) \citep{Bisk2020}, \socialiqa{} (\textbf{83.2\%}) \citep{Sap2019SocialIC}, \winogrande{} (\textbf{86.6\%}) \citep{Sakaguchi2020WINOGRANDEAA}, \cycic{} (\textbf{94.0\%}),\footnote{The \cycic{} dataset and leaderboard are available at \url{https://leaderboard.allenai.org/cycic}.} as well as the popular \commonsenseqa{} dataset (\textbf{79.3\%}) \citep{talmor-etal-2019-commonsenseqa}. Beyond setting records with the full training sets, our ablations show \universalmodel{} also improves data efficiency for all training dataset sizes.

For reproducibility, we publicly release the \universalmodel{} model and code, all the experimental results, and the \rainbow{} leaderboard at \url{https://github.com/allenai/rainbow}.

\section{Cost Equivalent Curves}
\label{sec:equivalent-curves}

\Costequivalentcurve{}s show \textit{equivalent costs} between the single-task baseline and a new transfer-based approach. In this work, we define \emph{cost} as \emph{the number of training examples in the target dataset}. Intuitively, we want to measure how many examples the new approach needs to match the single-task baseline's performance as the amount of data varies.

Figure~\ref{fig:equivalent-curve} illustrates \costequivalentcurve{}s with \commonsenseqa{} as the target dataset. The $x$-axis shows the number of examples used by the single-task baseline, while the $y$-axis shows the examples from the target dataset used by the new multitask method. The curve is where they achieve the same performance. The numbers on top of the figure show the performance corresponding to the number of baseline examples from the $x$-axis. For example, with 4.9k examples, the baseline achieves 70\% accuracy. For any number of examples the baseline might use, we can see how many examples the new approach would require to match it. In Figure~\ref{fig:equivalent-curve}, to match the baseline's performance on ${\sim}$10k examples, multitasking with \rainbow{} requires about 5k, while multitasking with \glue{} requires more than 10k. Thus, \emph{lower is better}, with curves below the diagonal ($y=x$) indicating that the new method improves over the baseline.

The construction of \costequivalentcurve{}s makes one technical assumption: the relationship between performance and cost is continuous and strictly monotonic (i.e., increasing or decreasing). This assumption holds empirically for parameters, compute, and data \citep{kaplan2020scaling}. Thus, we can safely estimate each learning curve with isotonic regression \citep{barlow1972statistical}, then construct the cost equivalent curve by mapping each dataset size to the baseline performance, finding the matching performance on the new method's curve, and seeing how many examples are required.

\Costequivalentcurve{}s visualize how a new approach impacts the cost-benefit trade-off, i.e. examples required for a given performance. This reframes the goal from pushing up performance on a fixed-size benchmark to most efficiently solving the problem. While we focus on data efficiency in this work, the idea of \costequivalentcurve{}s can be applied to other definitions of cost as well (e.g., GPU compute).
\section{\rainbow{}}
\label{sec:datasets}

\begin{figure*}[t]
    \centering
    \includegraphics[width=0.9\textwidth]{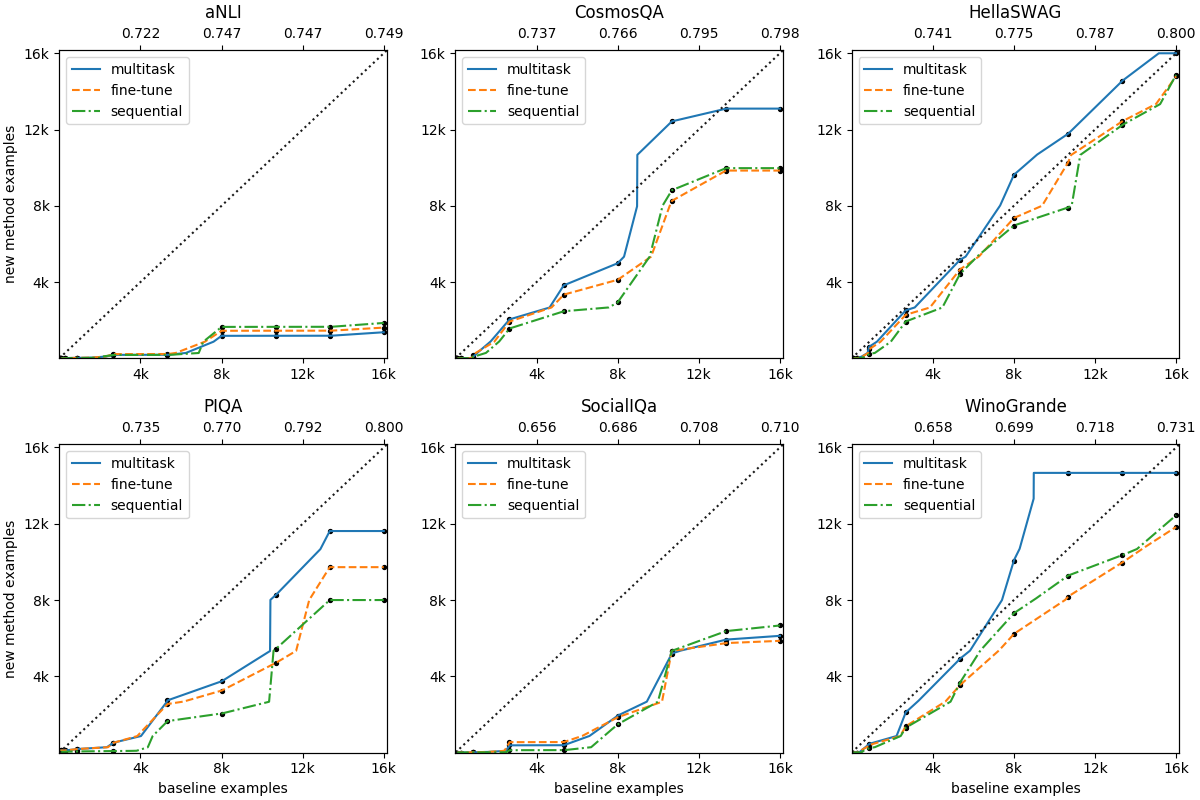}
    \caption{A comparison of transfer methods on \rainbow{} tasks with \tfive{-Large}. Each plot varies the data available for one task while using all data from the other five to generate the \costequivalentcurve{}. Performance is measured by dev set accuracy.}
    \label{fig:comparing-transfer}
\end{figure*}

\begin{table*}[t]
    \centering
    \begin{tabular}{lrrrrrr}
\toprule
\textbf{\textsc{\small{Transfer}}} & \textbf{\small{\anli{}}} & \textbf{\small{\cosmosqa{}}} &  \textbf{\small{\hellaswag{}}} &  \textbf{\small{\physicaliqa{}}} &  \textbf{\small{\socialiqa{}}} & \textbf{\small{\winogrande{}}} \\
\midrule
multitask           &          78.4 &          81.1 &          81.3 &          80.7 &          74.8 &          72.1 \\
fine-tune           &          79.2 &          82.6 & \textbf{83.1} & \textbf{82.2} &          75.2 &          78.2 \\
sequential          & \textbf{79.5} & \textbf{83.2} &          83.0 & \textbf{82.2} & \textbf{75.5} & \textbf{78.7} \\
\midrule
none                &          77.8 &          81.9 &          82.8 &          80.2 &          73.8 &          77.0 \\
\bottomrule
\end{tabular}

    \caption{A comparison of transfer methods' dev accuracy ($\%$) on the \rainbow{} tasks, using the \tfive{-Large} model.}
    \label{tab:comparing-transfer}
\end{table*}

We define \rainbow{}, a suite of commonsense benchmarks, with the following datasets. To keep evaluation clean-cut, we only chose multiple-choice question-answering datasets.

\begin{description}
\item{\textbf{\anli{}}} \citep{bhagavatula2020abductive} tests abductive reasoning in narratives. It asks models to identify the best explanation among several connecting a beginning and ending.
\item{\textbf{\cosmosqa{}}} \citep{Huang2019CosmosQM} asks commonsense reading comprehension questions about everyday narratives.
\item{\textbf{\hellaswag{}}} \citep{Zellers2019HellaSwagCA} requires models to choose the most plausible ending to a short context.
\item{\textbf{\piqa{}}} \citep{Bisk2020} is a multiple-choice question answering benchmark for physical commonsense reasoning.
\item{\textbf{\socialiqa{}}} \citep{Sap2019SocialIC} evaluates commonsense reasoning about social situations and interactions.
\item{\textbf{\winogrande{}}} \citep{Sakaguchi2020WINOGRANDEAA} is a large-scale collection of Winograd schema-inspired problems requiring reasoning about both social and physical interactions.
\end{description}

\section{Empirical Insights}
\label{sec:experiments}

We present results from our large-scale empirical study, using pretrained \tfive{-Large} to transfer between datasets. We've grouped our findings and their relevant figures around the four following thematic questions.


\begin{figure*}[t]
    \centering
    \includegraphics[width=0.9\textwidth]{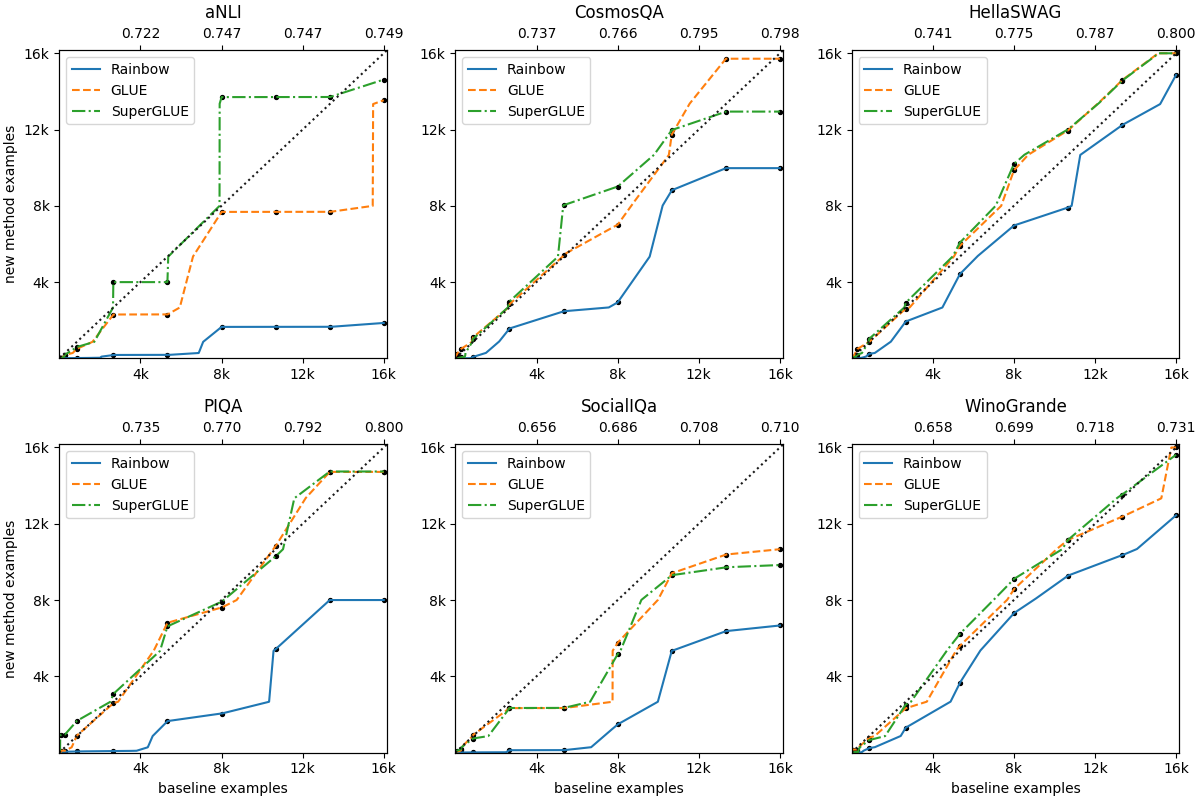}
    \caption{A comparison of multisets' transfer to \rainbow{} tasks using sequential training with \tfive{-Large}. Performance is measured by dev set accuracy. For transfer from \rainbow{}, we hold out the end task from the first round of fine-tuning.}
    \label{fig:comparing-multisets}
\end{figure*}

\begin{table*}[t]
    \centering
    \begin{tabular}{lrrrrrr}
\toprule
\textbf{\textsc{\small{Multiset}}} & \textbf{\small{\anli{}}} & \textbf{\small{\cosmosqa{}}} &  \textbf{\small{\hellaswag{}}} &  \textbf{\small{\piqa{}}} &  \textbf{\small{\socialiqa{}}} & \textbf{\small{\winogrande{}}} \\
\midrule
\glue{}      &          78.5 &          81.4 &          82.3 &          80.8 &          74.3 &          77.7 \\
\superglue{} &          79.1 &          82.2 &          82.5 &          80.7 &          74.6 &          77.6 \\
\rainbow{}   & \textbf{79.5} & \textbf{83.2} & \textbf{83.0} & \textbf{82.2} & \textbf{75.5} & \textbf{78.7} \\
\midrule
single task  &          77.8 &          81.9 &          82.8 &          80.2 &          73.8 &          77.0 \\
\bottomrule
\end{tabular}

    \caption{A comparison of dev accuracy for multisets' transfer to \rainbow{} via sequential training with \tfive{-Large}.}
    \label{tab:comparing-multisets}
\end{table*}


\begin{figure*}[t]
    \centering
    \includegraphics[width=0.9\textwidth]{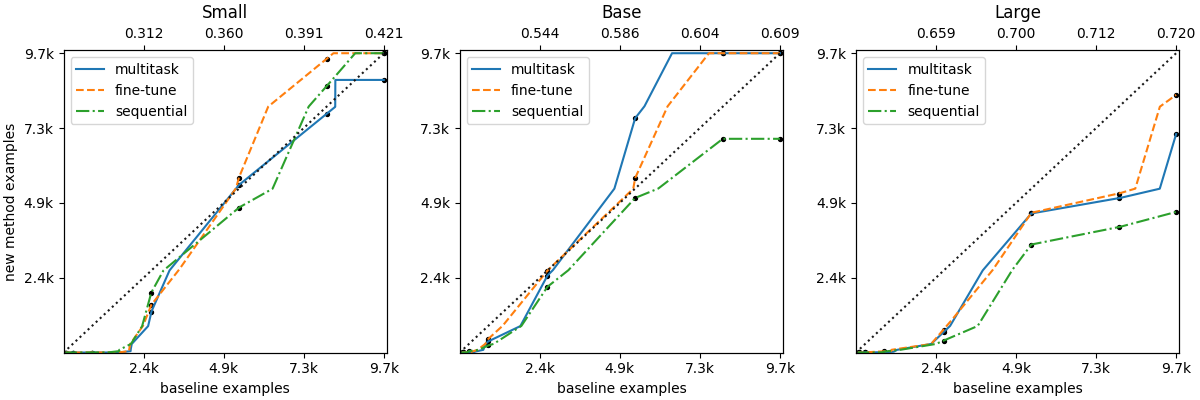}
    \caption{\Costequivalentcurve{}s comparing the effect of transfer across differently sized models on \commonsenseqa{}.}
    \label{fig:impact-of-size}
\end{figure*}

\subsection{What's the Best Approach for Transfer?}
\label{sec:experiments:comparing-transfer-methods}

We compare three recipes for intermediate-task transfer: 

\begin{description}
\item{\textbf{(1) multitask training}} \citep{caruana1995learningMany}: 
training on multiple datasets (\emph{including} the target dataset) all at once,
\item{\textbf{(2) sequential training}} \citep{Pratt1991DirectTO}:  
first training on multiple datasets (\emph{excluding} the target dataset) through multitask training, and then continuing to train on the target dataset alone,
\item{\textbf{(3) multitask fine-tuning}} \citep{liu-etal-2019-multi}:
first training on all datasets (\emph{including} the target dataset) through multitask training, and then continuing to fine-tune on the target dataset alone. 
\end{description}
Figure~\ref{fig:comparing-transfer} compares these three methods on each of the six \rainbow{} tasks, using the other five datasets for transfer.

\paragraph{Finding 1: Sequential training almost always matches or beats other approaches.} Generally, sequential and multitask fine-tune training use fewer examples to achieve the same performance as multitask training or the single task baseline.\footnote{Equivalently, they achieve better performance for the same number of examples.} For some tasks (\anli{} and \socialiqa{}), all three methods perform similarly; however, on the rest, sequential and multitask fine-tune training greatly improve data efficiency. While sequential and multitask fine-tune training are often comparable, sequential training appears to be slightly more data efficient, both from comparing \costequivalentcurve{}s in Figure~\ref{fig:comparing-transfer} and full dataset performance in Table~\ref{tab:comparing-transfer}.

\paragraph{Finding 2: Sequential training rarely hurts performance.} While multitask training doesn't always beat the single task baseline, sequential and multitask fine-tune training uniformly outperform it---for all \rainbow{} tasks and dataset sizes (including full datasets). This pattern mostly holds with other source and target tasks, especially for sequential training which rarely significantly harms performance.

\paragraph{Finding 3: Multitask training helps most often in the low-data regime.} One mystery researchers currently face is the inconsistent effect of multitask learning: sometimes it helps, sometimes it hurts, sometimes it has no effect. \Costequivalentcurve{}s reveal one potential explanation: multitask learning tends to help when data is scarce, but may hurt performance if data is plentiful. In Figure~\ref{fig:comparing-transfer}, all \costequivalentcurve{}s initially require fewer examples than the single-task baseline (the $y=x$ line), while on some tasks (\hellaswag{} and \winogrande{}) multitasking eventually needs more data than the baseline. Table~\ref{tab:comparing-transfer} reinforces this story, where multitask learning hurts performance on three of the six tasks (\cosmosqa{}, \hellaswag{}, and \winogrande{}), with \winogrande{} dropping from 77.0\% to 72.1\% accuracy. The fact that such trends depend on things like data size shows the importance of examining a range of scenarios: changing the context can even reverse one's conclusions.

\subsection{What Transfers Best for Common Sense?}
\label{sec:experiments:comparing-multisets}

Understanding when datasets transfer well is still an open and active area of research \citep{vu2020exploring, pruksachatkun2020intermediate}. At present, modelers usually pick datasets that seem similar to the target, whether due to format, domain, or something else. To investigate common sense transfer, we compare how the \rainbow{} tasks transfer to each other against two other popular dataset collections: \glue{} and \superglue{}. Following the insights from Section~\ref{sec:experiments:comparing-transfer-methods}, we use the strongest transfer method, sequential training, for the comparison. Figure~\ref{fig:comparing-multisets} presents \costequivalentcurve{}s and Table~\ref{tab:comparing-multisets} provides full dataset numbers.

\paragraph{Finding 4: \rainbow{} transfers best for common sense.} Across all six \rainbow{} tasks and all training set sizes, the \rainbow{} tasks transfer better to each other than \glue{} and \superglue{} do to them. The same result also holds for the popular benchmark \commonsenseqa{} when multitask training (Figure~\ref{fig:equivalent-curve}); though, when multitasking with JOCI \citep{zhang-etal-2017-ordinal}, an ordinal commonsense variant of natural language inference, \rainbow{} appears either not to help or to slightly hurt data efficiency---potentially more so than \glue{} and \superglue{}.\footnote{For these additional experiments, see the extended experimental results at \url{https://github.com/allenai/rainbow}.}

\paragraph{Finding 5: Only \rainbow{} uniformly beats the baseline.} With sequential training and \tfive{-Base} or larger, \rainbow{} improves data efficiency and performance for \textit{every} task considered. Importantly, this pattern breaks down when multitask training, for which no multiset uniformly improved performance. Thus, sequential training can unlock useful transfer even in contexts where multitask training cannot. Likewise, smaller models demonstrated less transfer, as discussed further in Section~\ref{sec:experiments:impact-of-size}. Consequently, \tfive{-Small} (the smallest model) did not always benefit. In contrast to \rainbow{}, \glue{} and \superglue{} often had little effect or slightly decreased data efficiency.

\paragraph{Caveats about \glue{}, \superglue{}, and \tfive{}.} There's an important caveat to note about \tfive{}, the model used in our experiments, and its relationship to \glue{} and \superglue{}. The off-the-shelf \tfive{}'s weights come from multitask pretraining, where many tasks are mixed with a language modeling objective to learn a powerful initialization for the weights. In fact, both \glue{} and \superglue{} were mixed into the pretraining \citep{2019t5}. So, while \rainbow{} clearly improves data efficiency and performance, our experiments do not determine whether some of the benefit comes from the novelty of \rainbow{}'s knowledge to \tfive{}, as opposed to containing more general information than \glue{} and \superglue{}.


\begin{figure*}[t]
    \centering
    \includegraphics[width=0.83\textwidth]{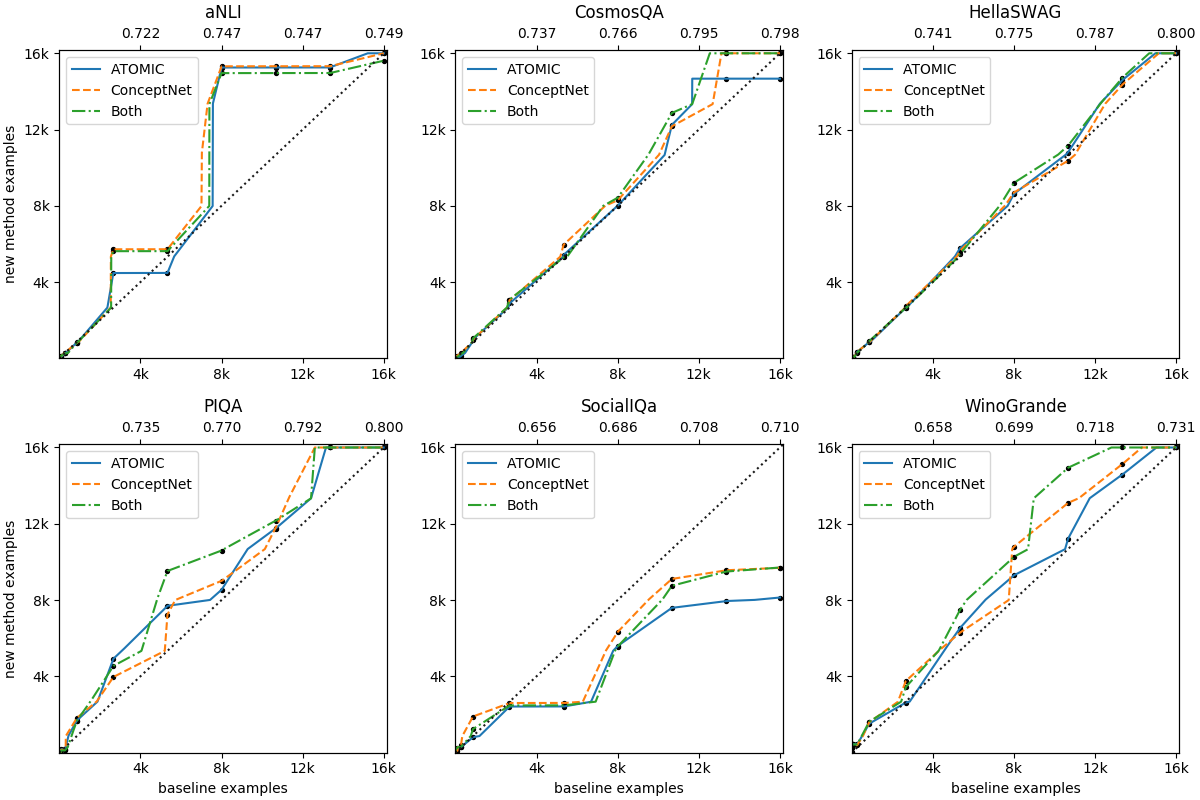}
    \caption{\Costequivalentcurve{}s comparing transfer from generative training on different common sense knowledge graphs using multitask training with \tfive{-Large}, across different \rainbow{} tasks. Performance is measured by dev set accuracy.}
    \label{fig:comparing-knowledge-graphs}
\end{figure*}

\begin{table*}[t]
    \centering
    \begin{tabular}{lrrrrrr}
\toprule
\textbf{\textsc{\small{Knowledge Graph}}} & \textbf{\small{\anli{}}} & \textbf{\small{\cosmosqa{}}} &  \textbf{\small{\hellaswag{}}} &  \textbf{\small{\piqa{}}} &  \textbf{\small{\socialiqa{}}} & \textbf{\small{\winogrande{}}} \\
\midrule
\atomic{}      & \textbf{78.3} &          81.8 & \textbf{82.8} &          79.9 & \textbf{75.0} & \textbf{78.2} \\
\conceptnet{}  &          78.0 &          81.8 &          82.5 &          80.5 &          74.3 &          76.3 \\
\dataset{Both} &          78.0 &          81.8 &          82.7 & \textbf{81.1} &          74.8 &          76.6 \\
\midrule
single task    &          77.8 & \textbf{81.9} & \textbf{82.8} &          80.2 &          73.8 &          77.0 \\
\bottomrule
\end{tabular}

    \caption{A comparison of dev accuracy when generatively training on knowledge graphs in a multitask setup using \tfive{-Large}.}
    \label{tab:comparing-knowledge-graphs}
\end{table*}

\subsection{Does Model Size Affect Transfer?}
\label{sec:experiments:impact-of-size}

Most of our exhaustive experiments use \tfive{-Large} (770M parameters), but in practice, we might prefer to use smaller models due to computational limitations. Thus, we investigate the impact of model size on intermediate-task transfer using the \tfive{-Base} (220M parameters) and \tfive{-Small} (60M parameters) models. Figure~\ref{fig:impact-of-size} presents the results for transferring with different model sizes from \rainbow{} to \commonsenseqa{}.

\paragraph{Finding 6: Larger models benefit more from transfer.}
Since larger pretrained models achieve substantially higher performance, it's difficult to compare transfer's effect across model size. The baselines start from very different places. \Costequivalentcurve{}s place everything in comparable units, \textit{equivalent baseline cost} (e.g., number of training examples). Capitalizing on this fact, Figure~\ref{fig:impact-of-size} compares transfer from \rainbow{} to \commonsenseqa{} across model size. The \costequivalentcurve{}s reveal a trend: larger models seem to benefit more from transfer, saving more examples over the relevant baselines. Since smaller models require more gradient updates to converge \citep{kaplan2020scaling}, it's important to note that we held the number of gradient updates fixed for comparison. Exploring whether this trend holds in different contexts, as well as theoretical explanations, are promising directions for future work.

\paragraph{Finding 7: Sequential training wins across model sizes.}
Figure~\ref{fig:impact-of-size} expands Finding~1, that sequential training generally matches or beats the other transfer approaches, by supporting it across model sizes. In all three plots, sequential training appears in line with or better than the other transfer methods.

\subsection{Can Models Transfer from Knowledge Graphs to QA Datasets?}
\label{sec:experiments:knowledge-graphs}

Due to reporting bias \citep{gordon2013reporting}, common sense rarely appears explicitly in text, though it does appear implicitly. While language models learn much of the common sense implicit in natural language \citep{trinh2018simple}, crowdsourced and expert curated knowledge might provide complementary information. To investigate, we used two popular common sense knowledge graphs, \conceptnet{} \citep{speer2017conceptnet} and \atomic{} \citep{sap2019atomic}, to create additional knowledge graph generation tasks \citep{Bosselut2019COMETCT}. In the forward direction, the model predicts the object given the subject and relation concatenated in XML tags. In the backward direction, the model predicts the subject given the object and relation. The results are summarized in Figure~\ref{fig:comparing-knowledge-graphs} and Table~\ref{tab:comparing-knowledge-graphs}.

\paragraph{Finding 8: Knowledge graph multitasking shows little impact.} The results are generally negative. Only \socialiqa{} benefits, which might come from the use of \atomic{} during its construction. We offer two possible explanations: the serialized language from the knowledge graphs is not in a QA format, and the knowledge graph completion task is generative while all other tasks are discriminative. These discrepancies may present too large an obstacle for effective transfer. Our findings encourage future research to better close the gap between knowledge graphs and datasets. Given sequential training's strength, as exemplified in Findings 1, 2, and 7, it may lead to different results than the multitask transfer we explore here.

\section{\universalmodel{}}
\label{sec:unicorn}

Finally, we present our \underline{uni}versal \underline{co}mmonsense \underline{r}easo\underline{n}ing model, \universalmodel{}. Motivated by Finding~1, our primary goal with \universalmodel{} is to provide a pretrained commonsense reasoning model ready to be fine-tuned on other downstream commonsense tasks. This is analogous to how off-the-shelf \tfive{} models are multitasked on NLP benchmarks such as \glue{} and \superglue{} as part of their pretraining.

In order to see the limit of the best performance achievable, we start by multitasking \tfive{-11B} on \rainbow{}. We then trained \universalmodel{} on each task individually, except for \winogrande{} which required separate handling since it evaluates models via a learning curve. For \winogrande{}, we multitasked the other five \rainbow{} datasets and then trained on \winogrande{}.\footnote{While sequential training for the \rainbow{} tasks would likely yield the best results, it would have required much more compute.} In each case, we used the same hyper-parameters as \universalmodel{} did during its initial multitask training, extending each of the 8 combinations tried at that stage. The best checkpoints were chosen using accuracy on dev.

\paragraph{SOTA on \rainbow{}.}
We establish new SOTA on all \rainbow{} datasets: \anli{} (\textbf{87.3\%}), \cosmosqa{} (\textbf{91.8\%}), \hellaswag{} (\textbf{93.9\%}), \physicaliqa{} (\textbf{90.1\%}), \socialiqa{} (\textbf{83.2\%}), and \winogrande{} (\textbf{86.6\%}).\footnote{All tasks use accuracy for evaluation except \winogrande{} which uses area under the dataset size--accuracy learning curve.}

\paragraph{SOTA on datasets beyond \rainbow{}.}
While SOTA results on \rainbow{} are encouraging, we still need to check if \universalmodel{}'s strong performance is confined to \rainbow{} or generalizes beyond it. Thus, we evaluated on two additional commonsense benchmarks: \cycic{} (\textbf{94.0\%}) and \commonsenseqa{} (\textbf{79.3\%}). Again, \universalmodel{} achieved SOTA on both.
\section{Related Work}
\label{sec:related-work}

\paragraph{Scaling Laws} In contemporary machine learning, simple methods that scale often outperform complex ones \citep{sutton_2019}. Accordingly, recent years have seen a sharp rise in compute used by state-of-the-art methods \citep{aiAndCompute2018}. Performance gains from increasing data, parameters, and training are not only reliable, but empirically predictable \citep{Hestness2017DeepLS, sun2017revisiting, rosenfeld2020a, kaplan2020scaling}. For example, \citet{sun2017revisiting} found that models need exponential data for improvements in accuracy.\footnote{Eventually, models saturate and need \textit{super-exponential} data.} These observations, that scaling is reliable, predictable, and critical to the current successes, motivate our focus on evaluation based on \textit{cost-benefit trade-offs}, i.e. the \costequivalentcurve{}.

\paragraph{Commonsense Benchmarks} Rapid progress in modeling has led to a major challenge for NLP: the creation of suitable benchmarks. Neural models often cue off statistical biases and annotation artifacts to solve datasets without understanding tasks \citep{gururangan-etal-2018-annotation}. To address this issue, recent commonsense benchmarks often use adversarial filtering \citep{zellers2018swagaf, Bras2020AdversarialFO}: a family of techniques that remove easily predicted examples from datasets. Besides \cosmosqa{}, all \rainbow{} tasks use this technique. Many more common sense benchmarks exist beyond what we could explore here \citep{roemmele2011choice, levesque2011winograd, mostafazadeh-etal-2016-corpus}.

\paragraph{Transfer Learning} Semi-supervised and transfer learning have grown into cornerstones of NLP. Early work learned unsupervised representations of words \citep{brown-etal-1992-class, mikolov2013distributedRepresentations}, while more recent work employs contextualized representations from neural language models \citep{peters-etal-2018-deep}. \citet{radford2018improving} demonstrated that language models could be fine-tuned directly to solve a wide-variety of tasks by providing the inputs encoded as text, while \citet{devlin-etal-2019-bert} and others improved upon the technique \citep{yang2019xlnet, liu2019roberta, lan2019albert}. Most relevant to this work, \citet{2019t5} introduced \tfive{} which built off previous work to reframe any NLP task as text-to-text, dispensing with the need for task-specific model adaptations.

\paragraph{Data Efficiency \& Evaluation} Other researchers have noted the importance of cost-benefit trade-offs in evaluation \citep{schwartz2019green}. \citet{dodge-etal-2019-show} advocate reporting the compute-performance trade-off caused by hyper-parameter tuning for new models, and provide an estimator for expected validation performance as a function of hyper-parameter evaluations. In an older work, \citet{clark1993using} evaluated the use of qualitative knowledge in terms of saved training examples, similarly to our \costequivalentcurve{}s. In contrast to our work, they fitted a linear trend to the learning curve and counted examples saved rather than plotting the numbers of examples that achieve equivalent performance.
\section{Conclusion}
\label{sec:conclusion}

Motivated by the fact that increased scale reliably improves performance for neural networks, we reevaluated existing techniques based on their data efficiency. To enable such comparisons, we introduced a new evaluation, the \costequivalentcurve{}, which improves over traditional learning curves by facilitating comparisons across otherwise hard-to-compare contexts. Our large-scale empirical study analyzed state-of-the-art techniques for transfer on pretrained language models, focusing on learning general, commonsense knowledge and evaluating on common sense tasks. In particular, we introduced a new collection of common sense datasets, \rainbow{}, and using the lessons from our empirical study trained a new model, \universalmodel{}, improving state-of-the-art results across 8 benchmarks. We hope others find our empirical study, new evaluation, \rainbow{}, and \universalmodel{} useful in their future work.
\section*{Acknowledgements}

We would like to thank the anonymous reviewers for their valuable feedback. This research was supported in part by NSF (IIS-1524371), the National Science Foundation Graduate Research Fellowship under Grant No. DGE 1256082, DARPA CwC through ARO (W911NF15-1- 0543), DARPA MCS program through NIWC Pacific (N66001-19-2-4031), and the Allen Institute for AI. Computations on \url{beaker.org} were supported in part by credits from Google Cloud. TPU machines for conducting experiments were generously provided by Google through the TensorFlow Research Cloud (TFRC) program.

{
  \small
  \bibliography{references}

\begin{thebibliography}{72}
\providecommand{\natexlab}[1]{#1}
\providecommand{\url}[1]{\texttt{#1}}
\providecommand{\urlprefix}{URL }
\expandafter\ifx\csname urlstyle\endcsname\relax
  \providecommand{\doi}[1]{doi:\discretionary{}{}{}#1}\else
  \providecommand{\doi}{doi:\discretionary{}{}{}\begingroup
  \urlstyle{rm}\Url}\fi

\bibitem[{Abadi et~al.(2016)Abadi, Barham, Chen, Chen, Davis, Dean, Devin,
  Ghemawat, Irving, Isard, Kudlur, Levenberg, Monga, Moore, Murray, Steiner,
  Tucker, Vasudevan, Warden, Wicke, Yu, and Zheng}]{abadi2016tensorflow}
Abadi, M.; Barham, P.; Chen, J.; Chen, Z.; Davis, A.; Dean, J.; Devin, M.;
  Ghemawat, S.; Irving, G.; Isard, M.; Kudlur, M.; Levenberg, J.; Monga, R.;
  Moore, S.; Murray, D.~G.; Steiner, B.; Tucker, P.; Vasudevan, V.; Warden, P.;
  Wicke, M.; Yu, Y.; and Zheng, X. 2016.
\newblock TensorFlow: A system for large-scale machine learning.
\newblock In \emph{12th USENIX Symposium on Operating Systems Design and
  Implementation (OSDI 16)}, 265--283.
\newblock
  \urlprefix\url{https://www.usenix.org/system/files/conference/osdi16/osdi16-abadi.pdf}.

\bibitem[{Agirre, M`arquez, and Wicentowski(2007)}]{agirre2007semantic}
Agirre, E.; M`arquez, L.; and Wicentowski, R., eds. 2007.
\newblock \emph{Proceedings of the Fourth International Workshop on Semantic
  Evaluations (SemEval-2007)}.
\newblock Prague, Czech Republic: Association for Computational Linguistics.

\bibitem[{Amodei and Hernandez(2018)}]{aiAndCompute2018}
Amodei, D.; and Hernandez, D. 2018.
\newblock AI and Compute.
\newblock \urlprefix\url{https://openai.com/blog/ai-and-compute/}.

\bibitem[{Bar~Haim et~al.(2006)Bar~Haim, Dagan, Dolan, Ferro, Giampiccolo,
  Magnini, and Szpektor}]{bar2006second}
Bar~Haim, R.; Dagan, I.; Dolan, B.; Ferro, L.; Giampiccolo, D.; Magnini, B.;
  and Szpektor, I. 2006.
\newblock The second {PASCAL} recognising textual entailment challenge .

\bibitem[{Barlow et~al.(1972)Barlow, Bartholomew, Bremner, and
  Brunk}]{barlow1972statistical}
Barlow, R.; Bartholomew, D.; Bremner, J.; and Brunk, H. 1972.
\newblock \emph{Statistical Inference Under Order Restrictions: The Theory and
  Application of Isotonic Regression}.
\newblock J. Wiley.
\newblock ISBN 9780471049708.

\bibitem[{Bentivogli et~al.(2009)Bentivogli, Dagan, Dang, Giampiccolo, and
  Magnini}]{bentivogli2009fifth}
Bentivogli, L.; Dagan, I.; Dang, H.~T.; Giampiccolo, D.; and Magnini, B. 2009.
\newblock The Fifth {PASCAL} Recognizing Textual Entailment Challenge .

\bibitem[{Bhagavatula et~al.(2020)Bhagavatula, {Le Bras}, Malaviya, Sakaguchi,
  Holtzman, Rashkin, Downey, Yih, and Choi}]{bhagavatula2020abductive}
Bhagavatula, C.; {Le Bras}, R.; Malaviya, C.; Sakaguchi, K.; Holtzman, A.;
  Rashkin, H.; Downey, D.; Yih, S. W.-t.; and Choi, Y. 2020.
\newblock Abductive commonsense reasoning.
\newblock \emph{ICLR} .

\bibitem[{Bisk et~al.(2020)Bisk, Zellers, {Le Bras}, Gao, and Choi}]{Bisk2020}
Bisk, Y.; Zellers, R.; {Le Bras}, R.; Gao, J.; and Choi, Y. 2020.
\newblock PIQA: Reasoning about Physical Commonsense in Natural Language.
\newblock In \emph{Thirty-Fourth AAAI Conference on Artificial Intelligence}.

\bibitem[{Bosselut et~al.(2019)Bosselut, Rashkin, Sap, Malaviya, Çelikyilmaz,
  and Choi}]{Bosselut2019COMETCT}
Bosselut, A.; Rashkin, H.; Sap, M.; Malaviya, C.; Çelikyilmaz, A.; and Choi,
  Y. 2019.
\newblock COMET: Commonsense Transformers for Automatic Knowledge Graph
  Construction.
\newblock In \emph{Proceedings of the 57th Annual Meeting of the Association
  for Computational Linguistics (ACL)}.

\bibitem[{Brown et~al.(1992)Brown, Della~Pietra, deSouza, Lai, and
  Mercer}]{brown-etal-1992-class}
Brown, P.~F.; Della~Pietra, V.~J.; deSouza, P.~V.; Lai, J.~C.; and Mercer,
  R.~L. 1992.
\newblock Class-Based \textit{n}-gram Models of Natural Language.
\newblock \emph{Computational Linguistics} 18(4): 467--480.
\newblock \urlprefix\url{https://www.aclweb.org/anthology/J92-4003}.

\bibitem[{Caruana(1995)}]{caruana1995learningMany}
Caruana, R. 1995.
\newblock Learning Many Related Tasks at the Same Time with Backpropagation.
\newblock In Tesauro, G.; Touretzky, D.~S.; and Leen, T.~K., eds.,
  \emph{Advances in Neural Information Processing Systems 7}, 657--664. MIT
  Press.
\newblock
  \urlprefix\url{http://papers.nips.cc/paper/959-learning-many-related-tasks-at-the-same-time-with-backpropagation.pdf}.

\bibitem[{Clark et~al.(2019)Clark, Lee, Chang, Kwiatkowski, Collins, and
  Toutanova}]{clark2019boolq}
Clark, C.; Lee, K.; Chang, M.-W.; Kwiatkowski, T.; Collins, M.; and Toutanova,
  K. 2019.
\newblock {B}ool{Q}: Exploring the Surprising Difficulty of Natural Yes/No
  Questions.
\newblock In \emph{Proceedings of NAACL-HLT 2019}.

\bibitem[{Clark and Matwin(1993)}]{clark1993using}
Clark, P.; and Matwin, S. 1993.
\newblock Using qualitative models to guide inductive learning.
\newblock In \emph{Proceedings of the 1993 international conference on machine
  learning}.

\bibitem[{Dagan, Glickman, and Magnini(2006)}]{dagan2006pascal}
Dagan, I.; Glickman, O.; and Magnini, B. 2006.
\newblock The {PASCAL} recognising textual entailment challenge.
\newblock In \emph{Machine learning challenges. evaluating predictive
  uncertainty, visual object classification, and recognising tectual
  entailment}, 177--190. Springer.

\bibitem[{De~Marneffe, Simons, and Tonhauser(2019)}]{demarneffe:cb}
De~Marneffe, M.-C.; Simons, M.; and Tonhauser, J. 2019.
\newblock {The CommitmentBank}: Investigating projection in naturally occurring
  discourse.
\newblock To appear in proceedings of Sinn und Bedeutung 23. Data can be found
  at https://github.com/mcdm/CommitmentBank/.

\bibitem[{Devlin et~al.(2019)Devlin, Chang, Lee, and
  Toutanova}]{devlin-etal-2019-bert}
Devlin, J.; Chang, M.-W.; Lee, K.; and Toutanova, K. 2019.
\newblock {BERT}: Pre-training of Deep Bidirectional Transformers for Language
  Understanding.
\newblock In \emph{Proceedings of the 2019 Conference of the North {A}merican
  Chapter of the Association for Computational Linguistics: Human Language
  Technologies, Volume 1 (Long and Short Papers)}, 4171--4186. Minneapolis,
  Minnesota: Association for Computational Linguistics.
\newblock \doi{10.18653/v1/N19-1423}.
\newblock \urlprefix\url{https://www.aclweb.org/anthology/N19-1423}.

\bibitem[{Dodge et~al.(2019)Dodge, Gururangan, Card, Schwartz, and
  Smith}]{dodge-etal-2019-show}
Dodge, J.; Gururangan, S.; Card, D.; Schwartz, R.; and Smith, N.~A. 2019.
\newblock Show Your Work: Improved Reporting of Experimental Results.
\newblock In \emph{Proceedings of the 2019 Conference on Empirical Methods in
  Natural Language Processing and the 9th International Joint Conference on
  Natural Language Processing (EMNLP-IJCNLP)}, 2185--2194. Hong Kong, China:
  Association for Computational Linguistics.
\newblock \doi{10.18653/v1/D19-1224}.
\newblock \urlprefix\url{https://www.aclweb.org/anthology/D19-1224}.

\bibitem[{Dolan and Brockett(2005)}]{dolan2005automatically}
Dolan, W.~B.; and Brockett, C. 2005.
\newblock Automatically constructing a corpus of sentential paraphrases.
\newblock In \emph{Proceedings of the International Workshop on Paraphrasing}.

\bibitem[{Giampiccolo et~al.(2007)Giampiccolo, Magnini, Dagan, and
  Dolan}]{giampiccolo2007third}
Giampiccolo, D.; Magnini, B.; Dagan, I.; and Dolan, B. 2007.
\newblock The third {PASCAL} recognizing textual entailment challenge.
\newblock In \emph{Proceedings of the ACL-PASCAL workshop on textual entailment
  and paraphrasing}, 1--9. Association for Computational Linguistics.

\bibitem[{Gordon and Van~Durme(2013)}]{gordon2013reporting}
Gordon, J.; and Van~Durme, B. 2013.
\newblock Reporting bias and knowledge acquisition.
\newblock In \emph{Proceedings of the 2013 workshop on Automated knowledge base
  construction}, 25--30. ACM.

\bibitem[{Gururangan et~al.(2018)Gururangan, Swayamdipta, Levy, Schwartz,
  Bowman, and Smith}]{gururangan-etal-2018-annotation}
Gururangan, S.; Swayamdipta, S.; Levy, O.; Schwartz, R.; Bowman, S.; and Smith,
  N.~A. 2018.
\newblock Annotation Artifacts in Natural Language Inference Data.
\newblock In \emph{NAACL}.
\newblock \urlprefix\url{https://www.aclweb.org/anthology/N18-2017/}.

\bibitem[{Hestness et~al.(2017)Hestness, Narang, Ardalani, Diamos, Jun,
  Kianinejad, Patwary, Yang, and Zhou}]{Hestness2017DeepLS}
Hestness, J.; Narang, S.; Ardalani, N.; Diamos, G.~F.; Jun, H.; Kianinejad, H.;
  Patwary, M. M.~A.; Yang, Y.; and Zhou, Y. 2017.
\newblock Deep Learning Scaling is Predictable, Empirically.
\newblock \emph{ArXiv} abs/1712.00409.

\bibitem[{Huang et~al.(2019)Huang, {Le Bras}, Bhagavatula, and
  Choi}]{Huang2019CosmosQM}
Huang, L.; {Le Bras}, R.; Bhagavatula, C.; and Choi, Y. 2019.
\newblock Cosmos QA: Machine Reading Comprehension with Contextual Commonsense
  Reasoning.
\newblock In \emph{EMNLP/IJCNLP}.

\bibitem[{Kaplan et~al.(2020)Kaplan, McCandlish, Henighan, Brown, Chess, Child,
  Gray, Radford, Wu, and Amodei}]{kaplan2020scaling}
Kaplan, J.; McCandlish, S.; Henighan, T.; Brown, T.~B.; Chess, B.; Child, R.;
  Gray, S.; Radford, A.; Wu, J.; and Amodei, D. 2020.
\newblock Scaling laws for neural language models.
\newblock \emph{arXiv preprint arXiv:2001.08361} .

\bibitem[{Khashabi et~al.(2018)Khashabi, Chaturvedi, Roth, Upadhyay, and
  Roth}]{khashabi2018looking}
Khashabi, D.; Chaturvedi, S.; Roth, M.; Upadhyay, S.; and Roth, D. 2018.
\newblock Looking beyond the surface: A challenge set for reading comprehension
  over multiple sentences.
\newblock In \emph{Proceedings of the 2018 Conference of the North American
  Chapter of the Association for Computational Linguistics: Human Language
  Technologies, Volume 1 (Long Papers)}, 252--262.

\bibitem[{Khashabi et~al.(2020)Khashabi, Min, Khot, Sabhwaral, Tafjord, Clark,
  and Hajishirzi}]{khashabi2020unifiedqa}
Khashabi, D.; Min, S.; Khot, T.; Sabhwaral, A.; Tafjord, O.; Clark, P.; and
  Hajishirzi, H. 2020.
\newblock UnifiedQA: Crossing Format Boundaries With a Single QA System.
\newblock \emph{arXiv preprint} .

\bibitem[{Lan et~al.(2019)Lan, Chen, Goodman, Gimpel, Sharma, and
  Soricut}]{lan2019albert}
Lan, Z.; Chen, M.; Goodman, S.; Gimpel, K.; Sharma, P.; and Soricut, R. 2019.
\newblock Albert: A lite bert for self-supervised learning of language
  representations.
\newblock \emph{arXiv preprint arXiv:1909.11942} .

\bibitem[{{Le Bras} et~al.(2020){Le Bras}, Swayamdipta, Bhagavatula, Zellers,
  Peters, Sabharwal, and Choi}]{Bras2020AdversarialFO}
{Le Bras}, R.; Swayamdipta, S.; Bhagavatula, C.; Zellers, R.; Peters, M.~E.;
  Sabharwal, A.; and Choi, Y. 2020.
\newblock Adversarial Filters of Dataset Biases.
\newblock \emph{ArXiv} abs/2002.04108.

\bibitem[{Levesque, Davis, and Morgenstern(2011)}]{levesque2011winograd}
Levesque, H.~J.; Davis, E.; and Morgenstern, L. 2011.
\newblock The {W}inograd schema challenge.
\newblock In \emph{{AAAI} Spring Symposium: Logical Formalizations of
  Commonsense Reasoning}, volume~46, 47.

\bibitem[{Liu et~al.(2019{\natexlab{a}})Liu, He, Chen, and
  Gao}]{liu-etal-2019-multi}
Liu, X.; He, P.; Chen, W.; and Gao, J. 2019{\natexlab{a}}.
\newblock Multi-Task Deep Neural Networks for Natural Language Understanding.
\newblock In \emph{Proceedings of the 57th Annual Meeting of the Association
  for Computational Linguistics}, 4487--4496. Florence, Italy: Association for
  Computational Linguistics.
\newblock \doi{10.18653/v1/P19-1441}.
\newblock \urlprefix\url{https://www.aclweb.org/anthology/P19-1441}.

\bibitem[{Liu et~al.(2019{\natexlab{b}})Liu, Ott, Goyal, Du, Joshi, Chen, Levy,
  Lewis, Zettlemoyer, and Stoyanov}]{liu2019roberta}
Liu, Y.; Ott, M.; Goyal, N.; Du, J.; Joshi, M.; Chen, D.; Levy, O.; Lewis, M.;
  Zettlemoyer, L.; and Stoyanov, V. 2019{\natexlab{b}}.
\newblock Roberta: A robustly optimized bert pretraining approach.
\newblock \emph{arXiv preprint arXiv:1907.11692} .

\bibitem[{Ma et~al.(2019)Ma, Francis, Lu, Nyberg, and
  Oltramari}]{ma-etal-2019-towards}
Ma, K.; Francis, J.; Lu, Q.; Nyberg, E.; and Oltramari, A. 2019.
\newblock Towards Generalizable Neuro-Symbolic Systems for Commonsense Question
  Answering.
\newblock In \emph{Proceedings of the First Workshop on Commonsense Inference
  in Natural Language Processing}, 22--32. Hong Kong, China: Association for
  Computational Linguistics.
\newblock \doi{10.18653/v1/D19-6003}.
\newblock \urlprefix\url{https://www.aclweb.org/anthology/D19-6003}.

\bibitem[{McCarthy(1959)}]{mccarthy59a}
McCarthy, J. 1959.
\newblock Programs with Common Sense.
\newblock In \emph{Proceedings of the {T}eddington Conference on the
  Mechanization of Thought Processes}, 75--91. London: Her Majesty's Stationary
  Office.

\bibitem[{Mikolov et~al.(2013)Mikolov, Sutskever, Chen, Corrado, and
  Dean}]{mikolov2013distributedRepresentations}
Mikolov, T.; Sutskever, I.; Chen, K.; Corrado, G.~S.; and Dean, J. 2013.
\newblock Distributed Representations of Words and Phrases and their
  Compositionality.
\newblock In Burges, C. J.~C.; Bottou, L.; Welling, M.; Ghahramani, Z.; and
  Weinberger, K.~Q., eds., \emph{Advances in Neural Information Processing
  Systems 26}, 3111--3119. Curran Associates, Inc.
\newblock
  \urlprefix\url{http://papers.nips.cc/paper/5021-distributed-representations-of-words-and-phrases-and-their-compositionality.pdf}.

\bibitem[{Mostafazadeh et~al.(2016)Mostafazadeh, Chambers, He, Parikh, Batra,
  Vanderwende, Kohli, and Allen}]{mostafazadeh-etal-2016-corpus}
Mostafazadeh, N.; Chambers, N.; He, X.; Parikh, D.; Batra, D.; Vanderwende, L.;
  Kohli, P.; and Allen, J. 2016.
\newblock A Corpus and Cloze Evaluation for Deeper Understanding of Commonsense
  Stories.
\newblock In \emph{Proceedings of the 2016 Conference of the North {A}merican
  Chapter of the Association for Computational Linguistics: Human Language
  Technologies}, 839--849. San Diego, California: Association for Computational
  Linguistics.
\newblock \doi{10.18653/v1/N16-1098}.
\newblock \urlprefix\url{https://www.aclweb.org/anthology/N16-1098}.

\bibitem[{Pedregosa et~al.(2011)Pedregosa, Varoquaux, Gramfort, Michel,
  Thirion, Grisel, Blondel, Prettenhofer, Weiss, Dubourg
  et~al.}]{pedregosa2011scikit}
Pedregosa, F.; Varoquaux, G.; Gramfort, A.; Michel, V.; Thirion, B.; Grisel,
  O.; Blondel, M.; Prettenhofer, P.; Weiss, R.; Dubourg, V.; et~al. 2011.
\newblock Scikit-learn: Machine learning in Python.
\newblock \emph{the Journal of machine Learning research} 12: 2825--2830.

\bibitem[{Peters et~al.(2018)Peters, Neumann, Iyyer, Gardner, Clark, Lee, and
  Zettlemoyer}]{peters-etal-2018-deep}
Peters, M.; Neumann, M.; Iyyer, M.; Gardner, M.; Clark, C.; Lee, K.; and
  Zettlemoyer, L. 2018.
\newblock Deep Contextualized Word Representations.
\newblock In \emph{Proceedings of the 2018 Conference of the North {A}merican
  Chapter of the Association for Computational Linguistics: Human Language
  Technologies, Volume 1 (Long Papers)}, 2227--2237. New Orleans, Louisiana:
  Association for Computational Linguistics.
\newblock \doi{10.18653/v1/N18-1202}.
\newblock \urlprefix\url{https://www.aclweb.org/anthology/N18-1202}.

\bibitem[{Pilehvar and Camacho-Collados(2019)}]{pilehvar2018wic}
Pilehvar, M.~T.; and Camacho-Collados, J. 2019.
\newblock {WiC}: The Word-in-Context Dataset for Evaluating Context-Sensitive
  Meaning Representations.
\newblock In \emph{Proceedings of NAACL-HLT}.

\bibitem[{Poliak et~al.(2018)Poliak, Haldar, Rudinger, Hu, Pavlick, White, and
  {Van Durme}}]{poliak2018dnc}
Poliak, A.; Haldar, A.; Rudinger, R.; Hu, J.~E.; Pavlick, E.; White, A.~S.; and
  {Van Durme}, B. 2018.
\newblock Collecting Diverse Natural Language Inference Problems for Sentence
  Representation Evaluation.
\newblock In \emph{Proceedings of EMNLP}.

\bibitem[{Pratt, Mostow, and Kamm(1991)}]{Pratt1991DirectTO}
Pratt, L.; Mostow, J.; and Kamm, C. 1991.
\newblock Direct Transfer of Learned Information Among Neural Networks.
\newblock In \emph{AAAI}.

\bibitem[{Pruksachatkun et~al.(2020)Pruksachatkun, Phang, Liu, Htut, Zhang,
  Pang, Vania, Kann, and Bowman}]{pruksachatkun2020intermediate}
Pruksachatkun, Y.; Phang, J.; Liu, H.; Htut, P.~M.; Zhang, X.; Pang, R.~Y.;
  Vania, C.; Kann, K.; and Bowman, S.~R. 2020.
\newblock Intermediate-Task Transfer Learning with Pretrained Models for
  Natural Language Understanding: When and Why Does It Work?
\newblock \emph{arXiv preprint arXiv:2005.00628} .

\bibitem[{Radford et~al.(2018)Radford, Narasimhan, Salimans, and
  Sutskever}]{radford2018improving}
Radford, A.; Narasimhan, K.; Salimans, T.; and Sutskever, I. 2018.
\newblock Improving language understanding by generative pre-training.
\newblock
  \urlprefix\url{https://s3-us-west-2.amazonaws.com/openai-assets/research-covers/language-unsupervised/language_understanding_paper.pdf}.

\bibitem[{Raffel et~al.(2019)Raffel, Shazeer, Roberts, Lee, Narang, Matena,
  Zhou, Li, and Liu}]{2019t5}
Raffel, C.; Shazeer, N.; Roberts, A.; Lee, K.; Narang, S.; Matena, M.; Zhou,
  Y.; Li, W.; and Liu, P.~J. 2019.
\newblock Exploring the Limits of Transfer Learning with a Unified Text-to-Text
  Transformer.
\newblock \emph{arXiv e-prints} .

\bibitem[{Rajpurkar et~al.(2016)Rajpurkar, Zhang, Lopyrev, and
  Liang}]{rajpurkar2016squad}
Rajpurkar, P.; Zhang, J.; Lopyrev, K.; and Liang, P. 2016.
\newblock {SQ}u{AD}: 100,000+ Questions for Machine Comprehension of Text.
\newblock In \emph{Proceedings of EMNLP}, 2383--2392. Association for
  Computational Linguistics.

\bibitem[{Roemmele, Bejan, and Gordon(2011)}]{roemmele2011choice}
Roemmele, M.; Bejan, C.~A.; and Gordon, A.~S. 2011.
\newblock Choice of plausible alternatives: An evaluation of commonsense causal
  reasoning.
\newblock In \emph{2011 AAAI Spring Symposium Series}.

\bibitem[{Rosenfeld et~al.(2020)Rosenfeld, Rosenfeld, Belinkov, and
  Shavit}]{rosenfeld2020a}
Rosenfeld, J.~S.; Rosenfeld, A.; Belinkov, Y.; and Shavit, N. 2020.
\newblock A Constructive Prediction of the Generalization Error Across Scales.
\newblock In \emph{International Conference on Learning Representations}.
\newblock \urlprefix\url{https://openreview.net/forum?id=ryenvpEKDr}.

\bibitem[{Rudinger et~al.(2018)Rudinger, Naradowsky, Leonard, and {Van
  Durme}}]{rudinger2018winogender}
Rudinger, R.; Naradowsky, J.; Leonard, B.; and {Van Durme}, B. 2018.
\newblock Gender Bias in Coreference Resolution.
\newblock In \emph{Proceedings of NAACL-HLT}.

\bibitem[{Sakaguchi et~al.(2020)Sakaguchi, {Le Bras}, Bhagavatula, and
  Choi}]{Sakaguchi2020WINOGRANDEAA}
Sakaguchi, K.; {Le Bras}, R.; Bhagavatula, C.; and Choi, Y. 2020.
\newblock WINOGRANDE: An Adversarial Winograd Schema Challenge at Scale.
\newblock In \emph{AAAI}.

\bibitem[{Sap et~al.(2019{\natexlab{a}})Sap, {Le Bras}, Allaway, Bhagavatula,
  Lourie, Rashkin, Roof, Smith, and Choi}]{sap2019atomic}
Sap, M.; {Le Bras}, R.; Allaway, E.; Bhagavatula, C.; Lourie, N.; Rashkin, H.;
  Roof, B.; Smith, N.~A.; and Choi, Y. 2019{\natexlab{a}}.
\newblock Atomic: An atlas of machine commonsense for if-then reasoning.
\newblock In \emph{Proceedings of the AAAI Conference on Artificial
  Intelligence}, volume~33, 3027--3035.

\bibitem[{Sap et~al.(2019{\natexlab{b}})Sap, Rashkin, Chen, {Le Bras}, and
  Choi}]{Sap2019SocialIC}
Sap, M.; Rashkin, H.; Chen, D.; {Le Bras}, R.; and Choi, Y. 2019{\natexlab{b}}.
\newblock Social IQA: Commonsense Reasoning about Social Interactions.
\newblock In \emph{EMNLP 2019}.

\bibitem[{Schwartz et~al.(2019)Schwartz, Dodge, Smith, and
  Etzioni}]{schwartz2019green}
Schwartz, R.; Dodge, J.; Smith, N.~A.; and Etzioni, O. 2019.
\newblock Green ai.
\newblock \emph{arXiv preprint arXiv:1907.10597} .

\bibitem[{Socher et~al.(2013)Socher, Perelygin, Wu, Chuang, Manning, Ng, and
  Potts}]{socher2013recursive}
Socher, R.; Perelygin, A.; Wu, J.; Chuang, J.; Manning, C.~D.; Ng, A.; and
  Potts, C. 2013.
\newblock Recursive deep models for semantic compositionality over a sentiment
  treebank.
\newblock In \emph{Proceedings of EMNLP}, 1631--1642.

\bibitem[{Speer, Chin, and Havasi(2017)}]{speer2017conceptnet}
Speer, R.; Chin, J.; and Havasi, C. 2017.
\newblock ConceptNet 5.5: An Open Multilingual Graph of General Knowledge.
\newblock In \emph{Proceedings of the Thirty-First AAAI Conference on
  Artificial Intelligence}, AAAI'17, 4444–4451. AAAI Press.

\bibitem[{Sun et~al.(2017)Sun, Shrivastava, Singh, and
  Gupta}]{sun2017revisiting}
Sun, C.; Shrivastava, A.; Singh, S.; and Gupta, A. 2017.
\newblock Revisiting unreasonable effectiveness of data in deep learning era.
\newblock In \emph{Proceedings of the IEEE international conference on computer
  vision}, 843--852.

\bibitem[{Sutton(2019)}]{sutton_2019}
Sutton, R.~S. 2019.
\newblock The Bitter Lesson.
\newblock
  \urlprefix\url{http://incompleteideas.net/IncIdeas/BitterLesson.html}.

\bibitem[{Talmor et~al.(2019)Talmor, Herzig, Lourie, and
  Berant}]{talmor-etal-2019-commonsenseqa}
Talmor, A.; Herzig, J.; Lourie, N.; and Berant, J. 2019.
\newblock {C}ommonsense{QA}: A Question Answering Challenge Targeting
  Commonsense Knowledge.
\newblock In \emph{Proceedings of the 2019 Conference of the North {A}merican
  Chapter of the Association for Computational Linguistics: Human Language
  Technologies, Volume 1 (Long and Short Papers)}, 4149--4158. Minneapolis,
  Minnesota: Association for Computational Linguistics.
\newblock \doi{10.18653/v1/N19-1421}.
\newblock \urlprefix\url{https://www.aclweb.org/anthology/N19-1421}.

\bibitem[{Tange(2011)}]{Tange2011a}
Tange, O. 2011.
\newblock GNU Parallel - The Command-Line Power Tool.
\newblock \emph{;login: The USENIX Magazine} 36(1): 42--47.
\newblock \doi{10.5281/zenodo.16303}.
\newblock \urlprefix\url{http://www.gnu.org/s/parallel}.

\bibitem[{Trinh and Le(2018)}]{trinh2018simple}
Trinh, T.~H.; and Le, Q.~V. 2018.
\newblock A simple method for commonsense reasoning.
\newblock \emph{arXiv preprint arXiv:1806.02847} .

\bibitem[{Vaswani et~al.(2017)Vaswani, Shazeer, Parmar, Uszkoreit, Jones,
  Gomez, Kaiser, and Polosukhin}]{vaswani2017attention}
Vaswani, A.; Shazeer, N.; Parmar, N.; Uszkoreit, J.; Jones, L.; Gomez, A.~N.;
  Kaiser, {\L}.; and Polosukhin, I. 2017.
\newblock Attention is all you need.
\newblock In \emph{Advances in neural information processing systems},
  5998--6008.

\bibitem[{Vu et~al.(2020)Vu, Wang, Munkhdalai, Sordoni, Trischler,
  Mattarella-Micke, Maji, and Iyyer}]{vu2020exploring}
Vu, T.; Wang, T.; Munkhdalai, T.; Sordoni, A.; Trischler, A.; Mattarella-Micke,
  A.; Maji, S.; and Iyyer, M. 2020.
\newblock Exploring and predicting transferability across nlp tasks.
\newblock \emph{arXiv preprint arXiv:2005.00770} .

\bibitem[{Wang et~al.(2019{\natexlab{a}})Wang, Pruksachatkun, Nangia, Singh,
  Michael, Hill, Levy, and Bowman}]{wang2019superglue}
Wang, A.; Pruksachatkun, Y.; Nangia, N.; Singh, A.; Michael, J.; Hill, F.;
  Levy, O.; and Bowman, S.~R. 2019{\natexlab{a}}.
\newblock Super{GLUE}: A Stickier Benchmark for General-Purpose Language
  Understanding Systems.
\newblock \emph{arXiv preprint 1905.00537} .

\bibitem[{Wang et~al.(2019{\natexlab{b}})Wang, Singh, Michael, Hill, Levy, and
  Bowman}]{wang2019glue}
Wang, A.; Singh, A.; Michael, J.; Hill, F.; Levy, O.; and Bowman, S.~R.
  2019{\natexlab{b}}.
\newblock {GLUE}: A Multi-Task Benchmark and Analysis Platform for Natural
  Language Understanding.
\newblock In the Proceedings of ICLR.

\bibitem[{Warstadt, Singh, and Bowman(2018)}]{warstadt2018neural}
Warstadt, A.; Singh, A.; and Bowman, S.~R. 2018.
\newblock Neural Network Acceptability Judgments.
\newblock \emph{arXiv preprint 1805.12471} .

\bibitem[{Williams, Nangia, and Bowman(2018)}]{williams2018broad}
Williams, A.; Nangia, N.; and Bowman, S.~R. 2018.
\newblock A Broad-Coverage Challenge Corpus for Sentence Understanding through
  Inference.
\newblock In \emph{Proceedings of NAACL-HLT}.

\bibitem[{{Williams} and {Zipser}(1989)}]{williams1989aLearning}
{Williams}, R.~J.; and {Zipser}, D. 1989.
\newblock A Learning Algorithm for Continually Running Fully Recurrent Neural
  Networks.
\newblock \emph{Neural Computation} 1(2): 270--280.

\bibitem[{Winograd(1972)}]{winograd1972understanding}
Winograd, T. 1972.
\newblock Understanding natural language.
\newblock \emph{Cognitive psychology} 3(1): 1--191.

\bibitem[{Yang et~al.(2019)Yang, Dai, Yang, Carbonell, Salakhutdinov, and
  Le}]{yang2019xlnet}
Yang, Z.; Dai, Z.; Yang, Y.; Carbonell, J.; Salakhutdinov, R.~R.; and Le, Q.~V.
  2019.
\newblock Xlnet: Generalized autoregressive pretraining for language
  understanding.
\newblock In \emph{Advances in neural information processing systems},
  5753--5763.

\bibitem[{Zellers et~al.(2018)Zellers, Bisk, Schwartz, and
  Choi}]{zellers2018swagaf}
Zellers, R.; Bisk, Y.; Schwartz, R.; and Choi, Y. 2018.
\newblock SWAG: A Large-Scale Adversarial Dataset for Grounded Commonsense
  Inference.
\newblock In \emph{Proceedings of the 2018 Conference on Empirical Methods in
  Natural Language Processing (EMNLP)}.

\bibitem[{Zellers et~al.(2019)Zellers, Holtzman, Bisk, Farhadi, and
  Choi}]{Zellers2019HellaSwagCA}
Zellers, R.; Holtzman, A.; Bisk, Y.; Farhadi, A.; and Choi, Y. 2019.
\newblock HellaSwag: Can a Machine Really Finish Your Sentence?
\newblock In \emph{ACL}.

\bibitem[{Zhang et~al.(2018)Zhang, Liu, Liu, Gao, Duh, and
  Durme}]{zhang2018record}
Zhang, S.; Liu, X.; Liu, J.; Gao, J.; Duh, K.; and Durme, B.~V. 2018.
\newblock {ReCoRD}: Bridging the Gap between Human and Machine Commonsense
  Reading Comprehension.
\newblock \emph{arXiv preprint 1810.12885} .

\bibitem[{Zhang et~al.(2017)Zhang, Rudinger, Duh, and
  Van~Durme}]{zhang-etal-2017-ordinal}
Zhang, S.; Rudinger, R.; Duh, K.; and Van~Durme, B. 2017.
\newblock Ordinal Common-sense Inference.
\newblock \emph{Transactions of the Association for Computational Linguistics}
  5: 379--395.
\newblock \doi{10.1162/tacl_a_00068}.
\newblock \urlprefix\url{https://www.aclweb.org/anthology/Q17-1027}.

\bibitem[{Zhu et~al.(2020)Zhu, Pang, Lan, and Cheng}]{zhu2020l2r2}
Zhu, Y.; Pang, L.; Lan, Y.; and Cheng, X. 2020.
\newblock L2R²: Leveraging Ranking for Abductive Reasoning.
\newblock In \emph{Proceedings of the 43rd International ACM SIGIR Conference
  on Research and Development in Information Retrieval}, SIGIR '20.
\newblock \doi{10.1145/3397271.3401332}.
\newblock \urlprefix\url{https://doi.org/10.1145/3397271.3401332}.

\end{thebibliography}
}
\nocite{Tange2011a}

\clearpage
\appendix
\section{Cost Equivalent Curves}
\label{app:equivalent-curves}

Section~\ref{sec:equivalent-curves} discusses the intuitions, assumptions, and visualization of \costequivalentcurve{}s at a high level. This appendix provides additional discussion as well as technical details for implementing \costequivalentcurve{}s.

The aim of \costequivalentcurve{}s is to visualize how an innovation impacts a cost-benefit trade-off, in a compact and intuitive way. Since \costequivalentcurve{}s are more general than the use case explored in this work (dataset size / performance trade-offs), we'll introduce more general terminology for discussing them, borrowing from the experimental design literature. The \textit{control} is the baseline approach (e.g., single task training), while the \textit{treatment} is the new approach or innovation (e.g., multitask or sequential training). \textit{Benefit} is a quantitative measure of how good the outcome is, like accuracy, while \textit{cost} measures what we pay to get it, such as dataset size or even dollars. Thus, \costequivalentcurve{}s can visualize how sequential training (the treatment) reduces data usage (the cost) compared to single task training (the control) when trying to achieve high accuracy (the benefit). Similarly, \costequivalentcurve{}s could visualize how Gaussian process optimization reduces hyper-parameter evaluations compared to random search when trying to achieve low perplexity on a language modeling task.

To construct \costequivalentcurve{}s, the main assumption is that the cost and benefit have a continuous, strictly monotonic (most often increasing) relationship. For machine learning, this assumption is satisfied empirically when using measures like expected cross-entropy against parameters, data, and compute \citep{kaplan2020scaling}. Since the cost and benefit share a monotonic relationship, we estimate the cost-benefit trade-offs using isotonic regression \citep{barlow1972statistical}. Concretely, we test the control and the treatment at a bunch of different costs, and measure the benefit. Then, we fit a curve to the control's results, \(\hat{f}_c\), and a curve to the treatment's results, \(\hat{f}_t\). Since the \costequivalentcurve{}, \(\hat{g}\), maps the control costs to the treatment costs achieving the same benefit, we can estimate it as:
\[ \hat{g} = \hat{f}_t^{-1} \circ \hat{f}_c \]
That is, we compose the \textit{inverse} cost-benefit curve for the treatment with the cost-benefit curve for the control. The inverse is guaranteed to exist because we assumed that the cost-benefit trade-offs are strictly monotonic.

Our implementation uses isotonic regression as implemented in scikit-learn \citep{pedregosa2011scikit}. To estimate the inverse curve, we switch the inputs and the outputs in the regression. The code may be found at \url{https://github.com/allenai/rainbow}.
\section{Datasets}
\label{app:datasets}

Our empirical study investigates transferring common sense from multisets (dataset collections) to various end tasks. Section~\ref{sec:datasets} presented a new multiset, \rainbow{}, for common sense transfer. In this appendix, Appendix~\ref{app:datasets:tasks} describes each end task we evaluated, Appendix~\ref{app:datasets:multisets} expands on \rainbow{} and the other multisets we tried, and Appendix~\ref{app:datasets:knowledge-graphs} details the knowledge graphs we used.

\subsection{Tasks}
\label{app:datasets:tasks}

Six tasks, \anli{}, \cosmosqa{}, \hellaswag{}, \piqa{}, \socialiqa{}, and \winogrande{}, compose \rainbow{}, as discussed in Section~\ref{sec:datasets}. Our experiments also use all six of these datasets as end tasks. In addition, we evaluated on \commonsenseqa{}, \joci{}, and \cycic{}. Each dataset is described below:

\paragraph{\anli{}} \citep{bhagavatula2020abductive} challenges models to infer the best explanation\footnote{Also known as \textit{abductive reasoning}.} connecting the beginning and ending of a story. Concretely, \anli{} presents models with the first and last sentences of a three sentence story. The model must choose among two alternative middles based on which provides the most plausible explanation.

\paragraph{\cosmosqa{}} \citep{Huang2019CosmosQM} tests models' reading comprehension by asking them to read in-between the lines. Each example presents a short passage along with a question dealing with commonsense causes, effects, inferences and counterfactuals drawn from the passage. To solve the task, models must choose the best answer among four candidates.

\paragraph{\hellaswag{}} \citep{Zellers2019HellaSwagCA} takes a context sentence and generates multiple completions using a language model. The machine generated endings often break commonsense world understanding, making it easy for humans to distinguish them from the original ending. In addition, \hellaswag{} uses \textit{adversarial filtering} \citep{zellers2018swagaf} to select the three distractor endings only from among those difficult for models to detect.

\paragraph{\piqa{}} \citep{Bisk2020} probes models' physical commonsense knowledge through goal-oriented question answering problems. The questions often explore object affordances, presenting a goal (e.g., \textit{``How do I find something lost on a carpet?''}) and then offering two solutions (such as \textit{``Put a solid seal on the end of your vacuum and turn it on''} vs. \textit{``Put a hair net on the end of your vacuum and turn it on''}). Models choose the best solution to solve the problem.

\paragraph{\socialiqa{}} \citep{Sap2019SocialIC} leverages \atomic{} \citep{sap2019atomic} to crowdsource a three-way multiple-choice benchmark evaluating the social and emotional common sense possessed by models. Questions explore people's motivations and reactions in a variety of social situations.

\paragraph{\winogrande{}} \citep{Sakaguchi2020WINOGRANDEAA} takes inspiration from \textit{winograd schemas} \citep{winograd1972understanding, levesque2011winograd} to create a large-scale dataset of coreference resolution problems requiring both physical and social common sense. Each question presents a sentence with a blank where a pronoun might be and two options to fill it. The questions often come in pairs where a single word changes between them, flipping which option is correct.

\paragraph{\commonsenseqa{}} \citep{talmor-etal-2019-commonsenseqa} offers general, challenging, common sense questions in a multiple-choice format. By construction, each question requires fine-grained world knowledge to distinguish between highly similar concepts. In particular, \commonsenseqa{} crowdsources questions by presenting annotators with three related concepts drawn from \conceptnet{} \citep{speer2017conceptnet}. The annotators then create three questions, each picking out one of the concepts as the correct answer. To increase the dataset's difficulty, an additional distractor from \conceptnet{} as well as one authored by a human were added to each question, for a total of five options.

\paragraph{\cycic{}\footnote{The \cycic{} dataset and leaderboard may be found at \url{https://leaderboard.allenai.org/cycic}}} offers five-way multiple-choice questions that touch on both common sense reasoning and knowledge over topics such as arithmetic, logic, time, and locations.

\paragraph{\joci{}} \citep{zhang-etal-2017-ordinal} (JHU Ordinal Commonsense Inference) generalizes natural language inference (NLI) to \textit{likely} implications. Each problem presents a context followed by a hypothesis. In contrast to traditional NLI which explores hard, logical implications, \joci{} instead explores \textit{likely} inferences from the context. Thus, each example comes with an ordinal label of the likelihood: \textit{very likely}, \textit{likely}, \textit{plausible}, \textit{technically possible}, or \textit{impossible}. In contrast to \citet{zhang-etal-2017-ordinal}, we treat the task as five-way classification and evaluate it with accuracy in order to make it uniform with other end tasks we explore.

\subsection{Multisets}
\label{app:datasets:multisets}

In addition to \rainbow{}, we use two other multisets for transfer. All three are described below.

\paragraph{\glue{}} \citep{wang2019glue} measures natural language understanding by evaluating models on a suite of classification tasks. In particular, \glue{} contains tasks for linguistic acceptability \citep{warstadt2018neural}, sentiment analysis \citep{socher2013recursive}, paraphrase \citep{dolan2005automatically, agirre2007semantic}\footnote{For more on Quora Question Pairs see \url{https://www.quora.com/q/quoradata/First-Quora-Dataset-Release-Question-Pairs}.}, natural language inference (sometimes constructed from other datasets) \citep{williams2018broad, rajpurkar2016squad, dagan2006pascal, bar2006second, giampiccolo2007third, bentivogli2009fifth, levesque2011winograd}, and general diagnostics.

\paragraph{\superglue{}} \cite{wang2019superglue} provides a more challenging successor to \glue{}, measuring natural language understanding with a broader range of more complex tasks. Specifically, \superglue{} comprises tasks for identifying when speakers implicitly assert something \citep{demarneffe:cb}, determining cause-effect relationships \citep{roemmele2011choice}, reading comprehension \citep{khashabi2018looking, zhang2018record}, natural language inference \citep{dagan2006pascal, bar2006second, giampiccolo2007third, bentivogli2009fifth, poliak2018dnc}, word sense disambiguation \citep{pilehvar2018wic}, winograd schemas \citep{levesque2011winograd}, true-false question answering \citep{clark2019boolq}, and gender bias diagnostics \citep{rudinger2018winogender}.

\paragraph{\rainbow{}} combines the six common sense benchmarks as we proposed in Section~\ref{sec:datasets}: \anli{} \citep{bhagavatula2020abductive}, \cosmosqa{} \citep{Huang2019CosmosQM}, \hellaswag{} \citep{Zellers2019HellaSwagCA}, \piqa{} \citep{Bisk2020}, \socialiqa{} \citep{Sap2019SocialIC}, and \winogrande{} \citep{Sakaguchi2020WINOGRANDEAA}. These multiple-choice datasets each measure different aspects of common sense, from likely sequences of events, to instrumental knowledge in physical situations, to theory of mind and social common sense.

\subsection{Knowledge Graphs}
\label{app:datasets:knowledge-graphs}

In addition to multisets, we explored common sense transfer from the following knowledge graphs in Section~\ref{sec:experiments:knowledge-graphs}:

\paragraph{\conceptnet{}} \citep{speer2017conceptnet} combines both expert curated and crowdsourced knowledge from various sources into a graph of concepts and relations. A \textit{concept} is a short natural language word or phrase, such as \textit{``water''}. Connecting concepts, there's a commonly used set of canonical \textit{relations} like \textit{\textsc{AtLocation}}. For example, \conceptnet{} contains the triple: \textit{``water''} \textit{\textsc{AtLocation}} \textit{``river''}. \conceptnet{} contains a significant amount of information beyond common sense; however, the common sense subset tends to focus on knowledge about objects and things.

\paragraph{\atomic{}} \citep{sap2019atomic} offers a rich source of knowledge about the relationships between events and common sense inferences about them. \atomic{} connects events described in natural language using relations that express things like pre-conditions, post-conditions, and plausible inferences based on the event. For example, \atomic{} contains the triple: \textit{``PersonX makes PersonY's coffee''} \textit{\textsc{oReact}} \textit{``PersonY will be grateful''}, where \textit{\textsc{oReact}} denotes the patient's (PersonY's) reaction.
\section{Training and Evaluation}
\label{app:training-and-evaluation}

This appendix describes the technical details of our training and evaluation setup, to help reproduce our experiments.

\subsection{Model and Implementation}
\label{app:training-and-evaluation:model-and-implementation}

All of our experiments are run with the state-of-the-art \tfive{} model \citep{2019t5}. \tfive{} is a text-to-text model built on top of the transformer architecture \citep{vaswani2017attention}. It has an encoder-decoder structure and is pretrained using a combination of masked language modeling \citep{devlin-etal-2019-bert} and multitask training on a large collection of NLP datasets. As a text-to-text model, \tfive{} frames every NLP problem as mapping input text to output text. All structural information in the input is linearized into a sequence of text, similarly to \citet{radford2018improving}, and all output is generated as a string when making predictions. For training, \tfive{} uses \textit{teacher forcing} \citep{williams1989aLearning}, i.e. maximum likelihood; for testing, \tfive{} greedily decodes the generated text. Thus, for \tfive{} to solve a task, one must first apply some straightforward preprocessing to frame it as text-to-text. Appendix~\ref{app:training-and-evaluation:preprocessing} describes the preprocessing we performed in more detail. Lastly,  \tfive{} is available in several model sizes: small (60M parameters), base (220M parameters), large (770M parameters), 3B (3B parameters), and 11B (11B parameters). For more information on \tfive{} and its pretraining, see \citet{2019t5}.

Our experiments use the original implementation, code, and weights for \tfive{}, which are publicly available at \url{https://github.com/google-research/text-to-text-transfer-transformer}. Our code uses the original \tfive{} implementation unmodified, only extending it with our own dataset preprocessing, reading, and task mixing. For deep learning operations, the implementation uses tensorflow \citep{abadi2016tensorflow}. Our code is available at \url{https://github.com/allenai/rainbow}.

\subsection{Preprocessing}
\label{app:training-and-evaluation:preprocessing}

\begin{figure}[h]
    \centering
    \includegraphics[width=\columnwidth]{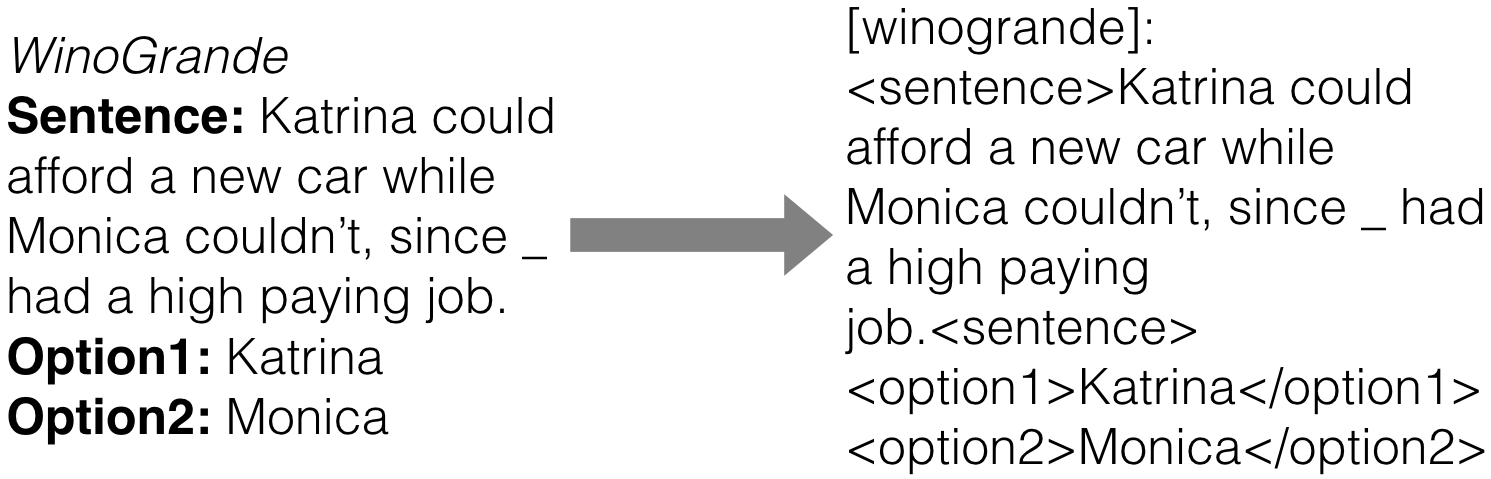}
    \caption{An example of the dataset preprocessing applied to an instance from \winogrande{}.}
    \label{fig:dataset-preprocessing}
\end{figure}

To model tasks as text-to-text, we need to convert their inputs and outputs into strings. Our preprocessing first prepends a string to each example signifying its dataset, e.g. \code{[socialiqa]:} for the \socialiqa{} task. Next, it wraps each feature in XML-like brackets with a unique tag identifying the feature, then joins them all together with newline characters. Figure~\ref{fig:dataset-preprocessing} depicts an example from \winogrande{}. Preprocessing for other tasks is similar.

\subsection{Training and Hyper-parameter Tuning}
\label{app:training-and-evaluation:training-and-hyper-parameter-tuning}

Following \citet{2019t5}, we converted all tasks to text-to-text and used teacher forcing \citep{williams1989aLearning} as the training objective, with greedy decoding for predictions. Our implementation reused the training and evaluation code from the original \tfive{} paper. For leaderboard submissions and test set evaluations, we built \universalmodel{} off of \tfive{-11B}. For all other experiments, we used \tfive{-Large} except when experiments specifically explore the impact of size, in which case the model size was explicitly indicated.

Hyper-parameters which were not set manually were tuned via grid search. In general, the fixed hyper-parameters and the grid used for search depended on the group of experiments, as outlined below. All hyper-parameters not mentioned were identical to those used in \citet{2019t5}.

\paragraph{Leaderboard Submissions} For leaderboard submissions and test set evaluations, \tfive{-11B} was initially multitasked on \rainbow{} with an equal task mixing rate for 25,000 gradient updates using different hyper-parameter combinations to produce the \universalmodel{}s. We then trained each on the end tasks separately for 25,000 gradient updates, saving a checkpoint every 2,500. The 10 most recent checkpoints were kept for early stopping, using dev set accuracy to choose the best checkpoint for evaluation. The grid search explored learning rates of 4e-3, 2e-3, 1e-3, and 5e-4 as well as batch sizes of 16 and 32.

\paragraph{Investigatory Experiments} Experiments which were not evaluated on the test set or submitted to a leaderboard used the \tfive{-Large} model as a starting point, unless explicitly noted otherwise (e.g., in experiments exploring the impact of model size). Training was carried out for 50,000 gradient updates, saving a checkpoint every 5,000 and keeping the 10 most recent. The batch size was fixed to 16. Grid search explored learning rates of 4e-3, 1e-3, and 2.5e-4. Depending on the specific experiment, other hyper-parameters were explored as well. For models trained on full datasets (rather than learning curves), we explored equal and dataset size-weighted mixing rates when multitasking. In sequential training, this meant that these rates were tried during the initial multitask training before training on the end task alone. For transferring knowledge graphs, we also explored predicting the subject-relation-object tuples in forward, backward, and bidirectional configurations. When producing learning curves, i.e. training the model on subsets of the full data, we used the equal mixing rate for all mixtures and the forward direction for knowledge graph transfer. Given the extensiveness of these experiments, we chose not to evaluate these models on the test sets to avoid test leakage; thus, reported results for these experiments are always the best score on dev.

For transfer techniques requiring two stages of training (i.e. multitask fine-tune and sequential training), we reused the hyper-parameters from the first stage of training in the second stage. For all tasks, we used accuracy as the evaluation metric.

To facilitate reproducibility and future research, we release results for all of our experiments, including hyper-parameter tuning. Download the results at \url{https://github.com/allenai/rainbow}. These tables contain all model evaluations and all hyper-parameter combinations tried in any given experiment.

\subsection{Hardware, Software, and Compute}
\label{app:training-and-evaluation:hardware-software-and-compute}

All experiments were run on Google Cloud using two Google Compute Engine virtual machine (VM) instances communicating with various TPUs. Experimental results were saved into Google Cloud Storage. Each VM had 20 vCPUs with 75GB of memory and ran Debian 9 (Stretch). One VM used Intel Skylake vCPUs while the other used Intel Haswell. Specific versions of libraries and other dependencies used are available and tracked in the code repository.

For hardware acceleration, we ran all the experiments using v3-8 TPUs when building off of \tfive{-Large} or smaller. For \tfive{-Small} and \tfive{-Large} we used a model parallelism of 8, while for \tfive{-Base} we used 4. The \tfive{-11B} models were trained using TPU v2-256 and v3-256s with a model parallelism of 16. Training times usually took several hours per run, so we ran many experiments in parallel on the VMs using GNU Parallel \citep{Tange2011a}.

\section{Leaderboards}
\label{app:leaderboards}

As discussed in Section~\ref{sec:unicorn}, \universalmodel{} achieves state-of-the-art performance across a number of popular commonsense benchmarks. This appendix collects those results along with the leaderboards' previous state-of-the-art and other useful baseline submissions for comparison.

\begin{table}[h!]
    \centering
    \anli{} \\
    \vspace{0.1cm}
    \begin{tabular}{lr}
        \toprule
        \textsc{Model} & \textsc{Accuracy} \\
        \midrule
        \model{BERT-Large} \citep{devlin-etal-2019-bert} & 66.8\% \\
        \model{RoBERTa-Large} \citep{liu2019roberta} & 83.9\% \\
        \model{L2R$^2$} \citep{zhu2020l2r2} & 86.8\% \\
        \universalmodel{} & \textbf{87.3\%} \\
        \midrule
        \textsc{Human} & 92.9\% \\
        \bottomrule
    \end{tabular}
    \caption{\anli{} leaderboard submissions.}
    \label{tab:anli-leaderboard}
\end{table}

\begin{table}[h!]
    \centering
    \cosmosqa{} \\
    \vspace{0.1cm}
    \begin{tabular}{lr}
        \toprule
        \textsc{Model} & \textsc{Accuracy} \\
        \midrule
        \model{RoBERTa-Large} \citep{liu2019roberta} & 83.5\% \\
        \model{ALBERT-xxLarge} \citep{lan2019albert} & 85.4\% \\
        \tfive{-11B} \citep{2019t5} & 90.3\% \\
        \universalmodel{} & \textbf{91.8\%} \\
        \midrule
        \textsc{Human} & 94.0\% \\
        \bottomrule
    \end{tabular}
    \caption{\cosmosqa{} leaderboard submissions.}
    \label{tab:cosmosqa-leaderboard}
\end{table}

\begin{table}[h!]
    \centering
    \hellaswag{} \\
    \vspace{0.1cm}
    \begin{tabular}{p{5.5cm}r}
        \toprule
        \textsc{Model} & \textsc{Accuracy} \\
        \midrule
        \model{RoBERTa-Large} \citep{liu2019roberta} & 81.7\% \\
        \model{HyKAS+CSKG} \citep{ma-etal-2019-towards} & 85.0\% \\
        \model{RoBERTa-Large Ensemble} \mbox{~~\citep{liu2019roberta}} & 85.5\% \\
        \universalmodel{} & \textbf{93.9\%} \\
        \midrule
        \textsc{Human} & 95.6\% \\
        \bottomrule
    \end{tabular}
    \caption{\hellaswag{} leaderboard submissions.}
    \label{tab:hellaswag-leaderboard}
\end{table}

\begin{table}[h!]
    \centering
    \physicaliqa{} \\
    \vspace{0.1cm}
    \begin{tabular}{lr}
        \toprule
        \textsc{Model} & \textsc{Accuracy} \\
        \midrule
        \model{BERT-Large} \citep{devlin-etal-2019-bert} & 66.7\% \\
        \model{RoBERTa-Large} \citep{liu2019roberta} & 79.4\% \\
        \model{UnifiedQA-3B} \citep{khashabi2020unifiedqa} & 85.3\% \\
        \universalmodel{} & \textbf{90.1\%} \\
        \midrule
        \textsc{Human} & 94.9\% \\
        \bottomrule
    \end{tabular}
    \caption{\physicaliqa{} leaderboard submissions.}
    \label{tab:physicaliqa-leaderboard}
\end{table}

\begin{table}[h!]
    \centering
    \socialiqa{} \\
    \vspace{0.1cm}
    \begin{tabular}{lr}
        \toprule
        \textsc{Model} & \textsc{Accuracy} \\
        \midrule
        \model{RoBERTa-Large} \citep{liu2019roberta} & 76.7\% \\
        \model{UnifiedQA-3B} \citep{khashabi2020unifiedqa} & 79.8\% \\
        \model{UGAmix} & 80.0\% \\
        \universalmodel{} & \textbf{83.2\%} \\
        \midrule
        \textsc{Human} & 88.1\% \\
        \bottomrule
    \end{tabular}
    \caption{\socialiqa{} leaderboard submissions.}
    \label{tab:socialiqa-leaderboard}
\end{table}

\begin{table}[h!]
    \centering
    \winogrande{} \\
    \vspace{0.1cm}
    \begin{tabular}{lr}
        \toprule
        \textsc{Model} & \textsc{AUC} \\
        \midrule
        \model{BERT-Large} \citep{devlin-etal-2019-bert} & 52.9\% \\
        \model{RoBERTa-Large} \citep{liu2019roberta} & 66.4\%\\
        \model{UnifiedQA-11B} \citep{khashabi2020unifiedqa} & 85.7\% \\
        \universalmodel{} & \textbf{86.6\%} \\
        \midrule
        \textsc{Human} & 94.0\% \\
        \bottomrule
    \end{tabular}
    \caption{\winogrande{} leaderboard submissions. \textit{AUC} is the area under the dataset-size vs. accuracy learning curve.}
    \label{tab:winogrande-leaderboard}
\end{table}

\begin{table}[h!]
    \centering
    \cycic{} \\
    \vspace{0.1cm}
    \begin{tabular}{lr}
        \toprule
        \textsc{Model} & \textsc{Accuracy} \\
        \midrule
        \model{RoBERTa-Large} \citep{liu2019roberta} & 91.3\%\\
        \model{PRv2} & 91.4\% \\
        \universalmodel{} & \textbf{94.0\%} \\
        \midrule
        \textsc{Human} & 90.0\% \\
        \bottomrule
    \end{tabular}
    \caption{\cycic{} leaderboard submissions.}
    \label{tab:cycic-leaderboard}
\end{table}

\begin{table}[h!]
    \centering
    \commonsenseqa{} \\
    \vspace{0.1cm}
    \begin{tabular}{lr}
        \toprule
        \textsc{Model} & \textsc{Accuracy} \\
        \midrule
        \model{RoBERTa-Large} \citep{liu2019roberta} & 72.1\% \\
        \tfive{-11B} \citep{2019t5} & 78.1\%\\
        \model{UnifiedQA-11B} \citep{khashabi2020unifiedqa} & 79.1\% \\
        \universalmodel{} & \textbf{79.3\%} \\
        \midrule
        \textsc{Human} & 88.9\% \\
        \bottomrule
    \end{tabular}
    \caption{\commonsenseqa{} leaderboard submissions.}
    \label{tab:commonsenseqa-leaderboard}
\end{table}

Since \commonsenseqa{} used \conceptnet{} in its construction, its authors have split leaderboard submissions into two categories: models that do and that do not use \conceptnet{}. Models using \conceptnet{} can gain an advantage by eliminating the human authored distractor options. \universalmodel{} holds the current state-of-the-art among models which do \textit{not} use \conceptnet{}. The state-of-the-art model using \conceptnet{} combines the knowledge graph with \model{ALBERT} \citep{lan2019albert}\footnote{For more, see \url{https://github.com/jessionlin/csqa/blob/master/Model_details.md}} and scores 79.5\% accuracy.

\paragraph{Hyper-parameters} For each of the submissions, we used the following hyper-parameters. \anli{} used a learning rate of 5e-4 and a batch size of 16. \cosmosqa{} used a learning rate of 2e-3 and a batch size of 32. \hellaswag{} used a learning rate of 2e-3 and a batch size of 32. \physicaliqa{} used a learning rate of 2e-3 and a batch size of 32. \socialiqa{} used a learning rate of 5e-4 and a batch size of 32. \winogrande{}-xs used a learning rate of 2e-3 and a batch size of 16, \winogrande{}-s used a learning rate of 2e-3 and a batch size of 16, \winogrande{}-m used a learning rate of 5e-4 and a batch size of 32, \winogrande{}-l used a learning rate of 1e-3 and a batch size of 16,
and \winogrande{}-xl used a learning rate of 2e-3 and a batch size of 16. \cycic{} had a learning rate of 5e-4 and a batch size of 32, while \commonsenseqa{} had a learning rate of 1e-3 and a batch size of 32.
\section{Experiments}
\label{app:experiments}

This appendix provides additional figures illustrating the findings as well as tables for all the experiments. In addition, the code used to run these experiments may be found at \url{https://github.com/allenai/rainbow}, and the models, experimental results (in CSV format) and even more figures, may be downloaded there as well.

\subsection{Transferring to the \rainbow{} Tasks}
\label{app:experiments:transferring-to-rainbow}

\begin{figure*}[h]
    \centering
    \includegraphics[width=0.85\textwidth]{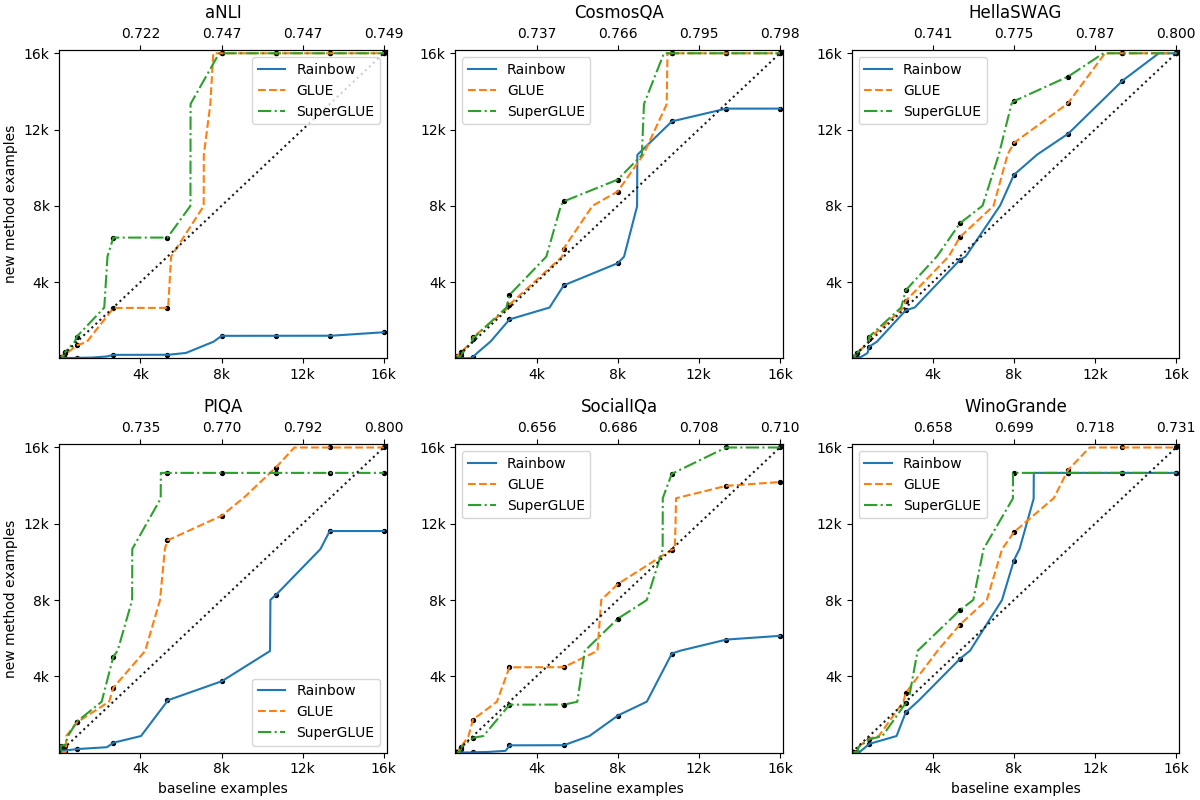}
    \caption{A comparison of multisets for transfer to the \rainbow{} tasks using multitask training with \tfive{-Large}. Performance is dev accuracy.}
    \label{fig:compare-multisets_multitask}
\end{figure*}

\begin{figure*}[h]
    \centering
    \includegraphics[width=0.85\textwidth]{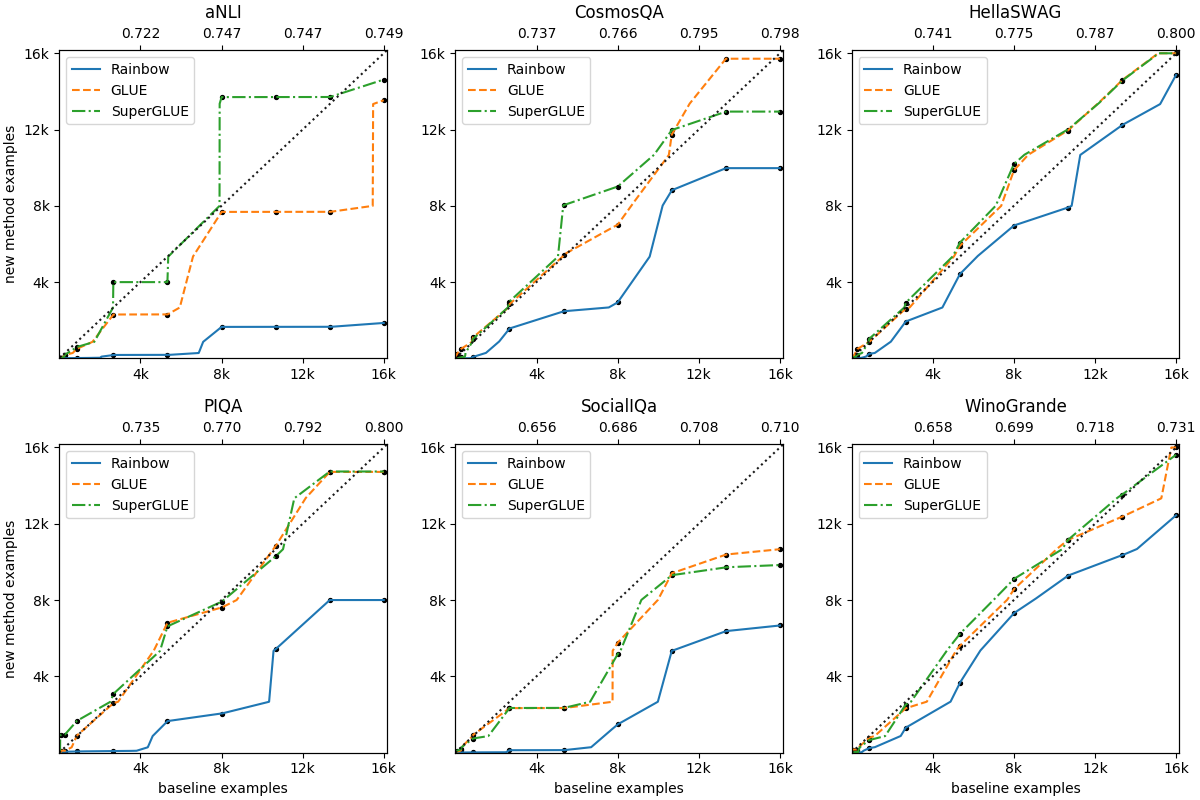}
    \caption{A comparison of multisets for transfer to the \rainbow{} tasks using sequential training with \tfive{-Large}. Performance is dev accuracy.}
    \label{fig:compare-multisets_sequential}
\end{figure*}

\begin{figure*}[h]
    \centering
    \includegraphics[width=0.85\textwidth]{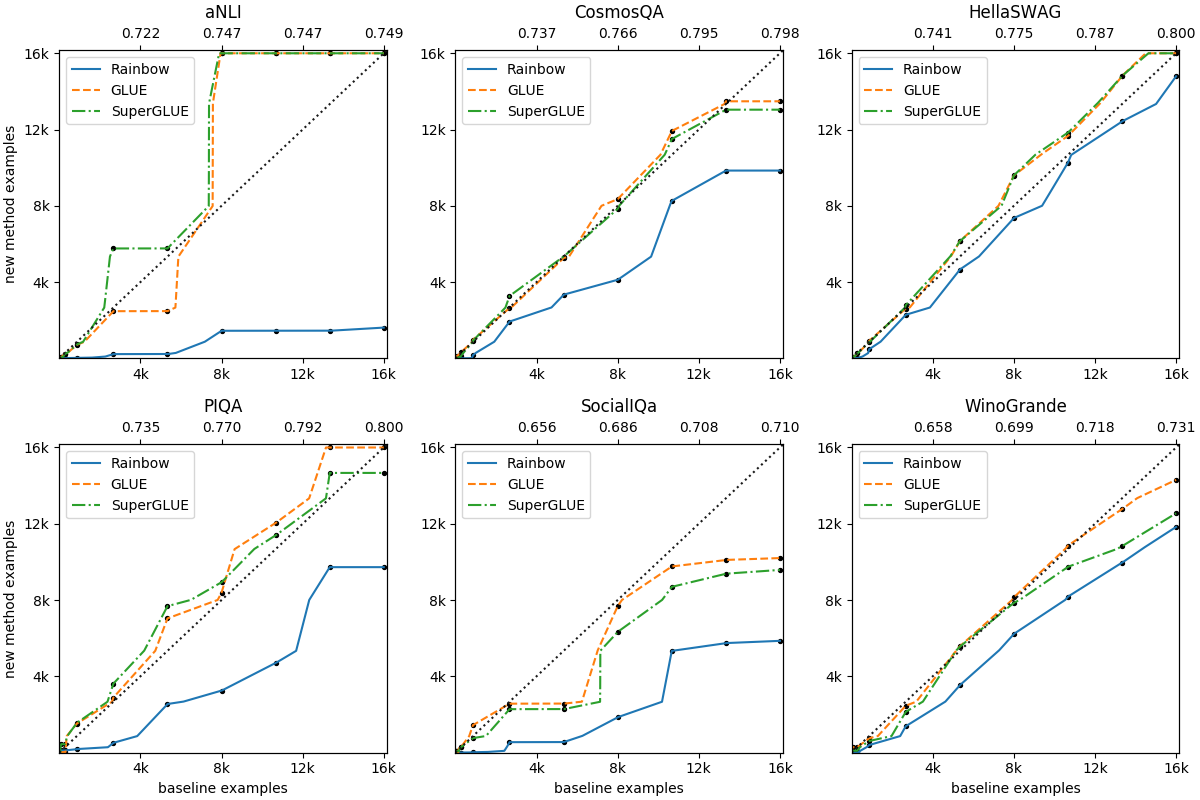}
    \caption{A comparison of multisets for transfer to the \rainbow{} tasks using multitask fine-tune training with \tfive{-Large}. Performance is dev accuracy.}
    \label{fig:compare-multisets_multitask-fine-tune}
\end{figure*}

\begin{figure*}[h]
    \centering
    \includegraphics[width=0.85\textwidth]{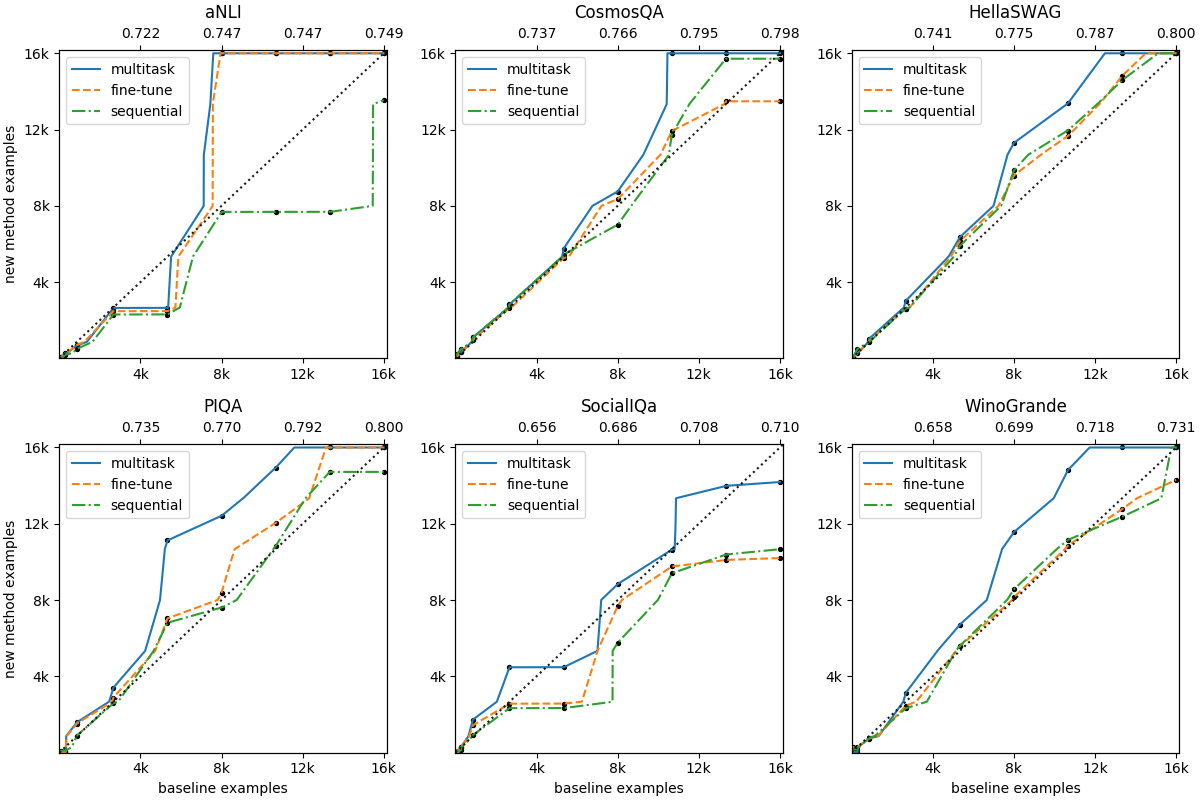}
    \caption{A comparison of methods for transferring \glue{} to the \rainbow{} tasks with \tfive{-Large}. Performance is dev accuracy.}
    \label{fig:compare-transfer_glue}
\end{figure*}

\begin{figure*}[h]
    \centering
    \includegraphics[width=0.85\textwidth]{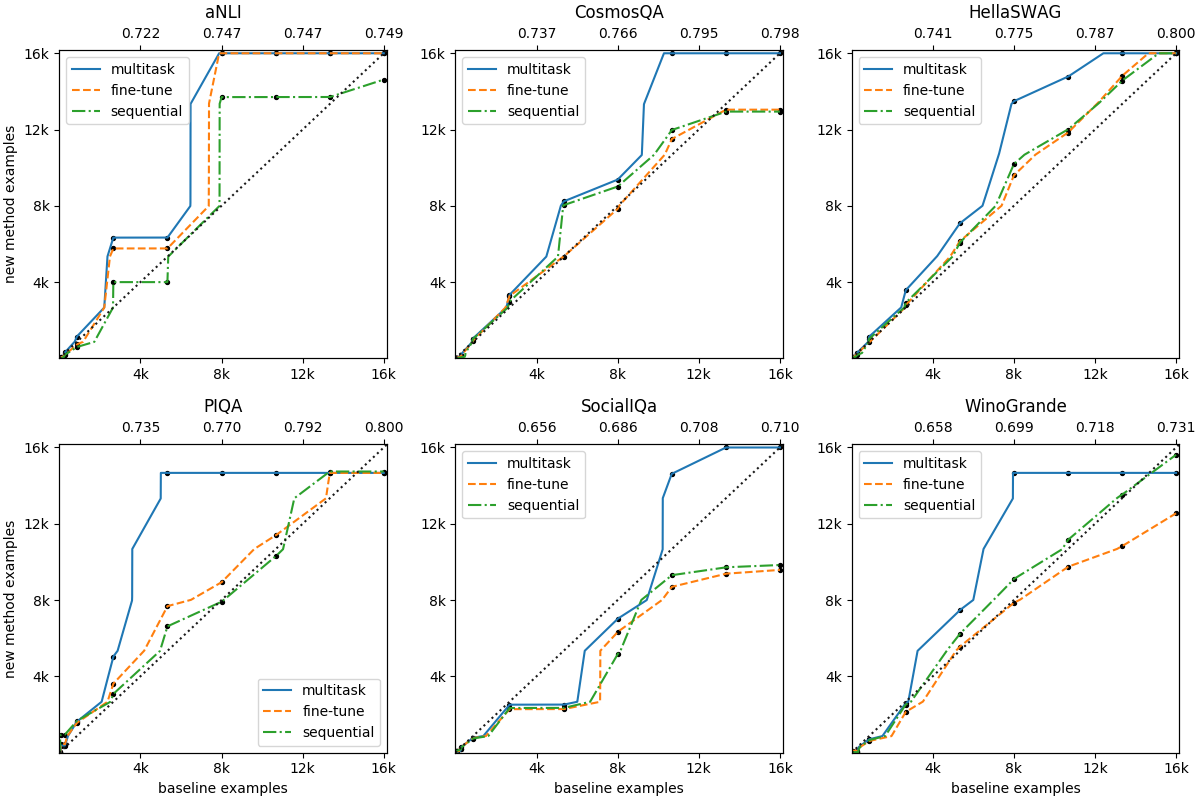}
    \caption{A comparison of methods for transferring \superglue{} to the \rainbow{} tasks with \tfive{-Large}. Performance is dev accuracy.}
    \label{fig:compare-transfer_super-glue}
\end{figure*}

\begin{figure*}[h]
    \centering
    \includegraphics[width=0.85\textwidth]{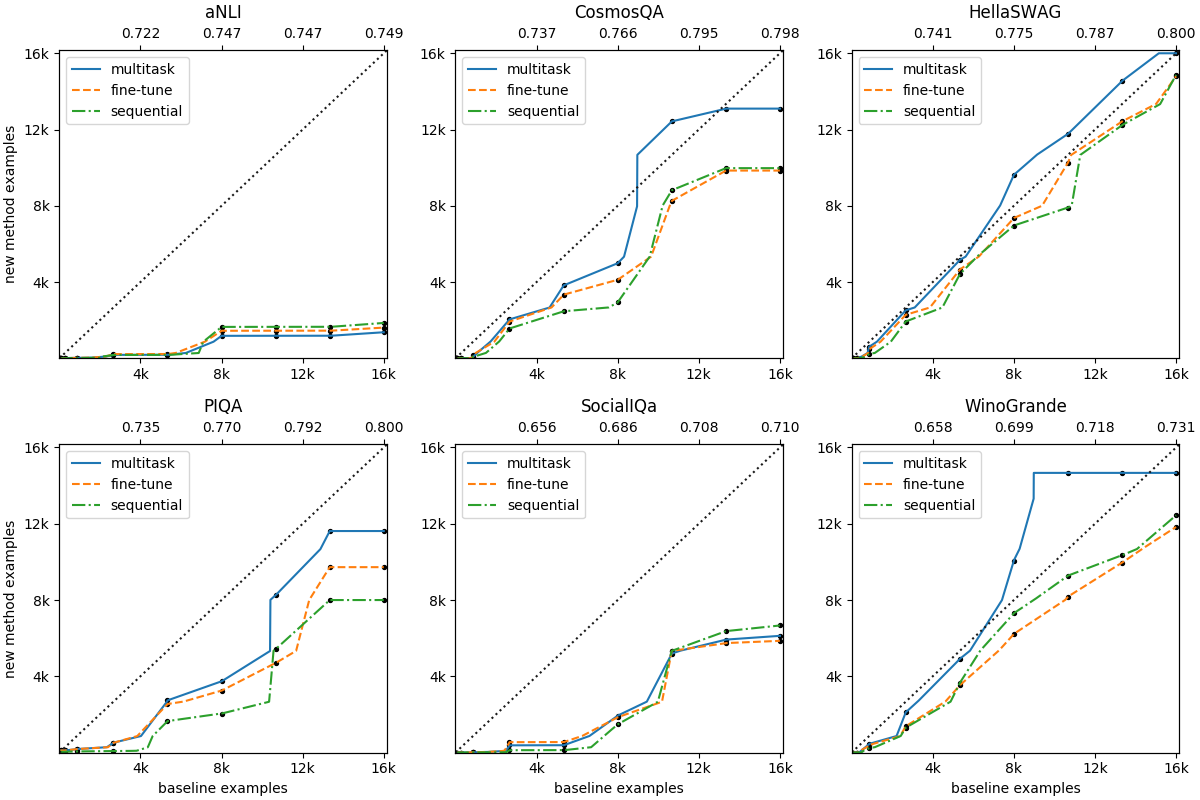}
    \caption{A comparison of methods for transferring the \rainbow{} tasks to the \rainbow{} tasks with \tfive{-Large}. Each plot treats its end task as held out from \rainbow{}, using the other five tasks for transfer. Performance is dev accuracy.}
    \label{fig:compare-transfer_rainbow}
\end{figure*}

These figures and tables use \rainbow{} for the end tasks:

\begin{description}
\item{\textbf{Figure~\ref{fig:compare-multisets_multitask}}} A comparison of different multisets using multitask training
\item{\textbf{Figure~\ref{fig:compare-multisets_sequential}}} A comparison of different multisets using sequential training
\item{\textbf{Figure~\ref{fig:compare-multisets_multitask-fine-tune}}} A comparison of different multisets using multitask fine-tune training
\item{\textbf{Figure~\ref{fig:compare-transfer_glue}}} A comparison of transfer methods on \glue{}
\item{\textbf{Figure~\ref{fig:compare-transfer_super-glue}}} A comparison of transfer methods on \superglue{}
\item{\textbf{Figure~\ref{fig:compare-transfer_rainbow}}} A comparison of transfer methods on \rainbow{}
\item{\textbf{Table~\ref{tab:single-task_full-tasks}}} Single task baselines using the full training data
\item{\textbf{Table~\ref{tab:multiset_full-tasks}}} The performance using transfer and the full training data
\item{\textbf{Table~\ref{tab:single-task_learning-curves}}} Single task learning curves
\item{\textbf{Table~\ref{tab:anli_learning-curves}}} \anli{} learning curves with transfer
\item{\textbf{Table~\ref{tab:cosmosqa_learning-curves}}} \cosmosqa{} learning curves with transfer
\item{\textbf{Table~\ref{tab:hellaswag_learning-curves}}} \hellaswag{} learning curves with transfer
\item{\textbf{Table~\ref{tab:physicaliqa_learning-curves}}} \physicaliqa{} learning curves with transfer
\item{\textbf{Table~\ref{tab:socialiqa_learning-curves}}} \socialiqa{} learning curves with transfer
\item{\textbf{Table~\ref{tab:winogrande_learning-curves}}} \winogrande{} learning curves with transfer
\end{description}


\begin{table*}
    \centering
    \begin{tabular}{lrrrrrr}
\toprule
 &\multicolumn{6}{c}{\small{Task}}\\ \cline{2-7}
model &  \small{\anli{}} &  \small{\cosmosqa{}} &  \small{\hellaswag{}} &  \small{\physicaliqa{}} &  \small{\socialiqa{}} &  \small{\winogrande{}} \\
\midrule
large &             77.8 &                 81.9 &                  82.8 &                    80.2 &                  73.8 &                   77.0 \\
\bottomrule
\end{tabular}

    \caption{Single task baselines using the full training data.}
    \label{tab:single-task_full-tasks}
\end{table*}

\begin{table*}
    \centering
    \begin{tabular}{llrrrrrr}
\toprule
 & &\multicolumn{6}{c}{\small{Task}}\\ \cline{3-8}
multiset & transfer &  \small{\anli{}} &  \small{\cosmosqa{}} &  \small{\hellaswag{}} &  \small{\physicaliqa{}} &  \small{\socialiqa{}} &  \small{\winogrande{}} \\
\midrule
\multirow{3}{*}{\glue{}} & fine-tune &             78.5 &                 82.4 &                  82.9 &                    80.1 &                  74.3 &                   78.4 \\
             & multitask &             77.2 &                 80.8 &                  81.8 &                    77.6 &                  74.7 &                   76.0 \\
             & sequential &             78.5 &                 81.4 &                  82.3 &                    80.8 &                  74.3 &                   77.9 \\
\cline{1-8}
\multirow{3}{*}{\rainbow{}} & fine-tune &             79.2 &                 82.6 &                  83.1 &                    82.2 &                  75.2 &                   78.2 \\
             & multitask &             78.4 &                 81.1 &                  81.3 &                    80.7 &                  74.8 &                   72.1 \\
             & sequential &             79.5 &                 83.2 &                  83.0 &                    82.2 &                  75.5 &                   78.7 \\
\cline{1-8}
\multirow{3}{*}{\superglue{}} & fine-tune &             78.5 &                 81.7 &                  82.8 &                    80.0 &                  74.7 &                   78.5 \\
             & multitask &             77.7 &                 78.9 &                  80.5 &                    70.7 &                  72.3 &                   69.8 \\
             & sequential &             79.1 &                 82.2 &                  82.5 &                    80.7 &                  74.6 &                   77.6 \\
\bottomrule
\end{tabular}

    \caption{The performance using transfer and the full training data.}
    \label{tab:multiset_full-tasks}
\end{table*}

\begin{table*}
    \centering
    \begin{tabular}{lrrrrrrrrrrrr}
\toprule
 &\multicolumn{12}{c}{\small{Size}}\\ \cline{2-13}
task           &  \small{4} &  \small{10} &  \small{30} &  \small{91} &  \small{280} &  \small{865} &  \small{2667} &  \small{5334} &  \small{8000} &  \small{10667} &  \small{13334} &  \small{16000} \\
\midrule
\anli{}        &       49.2 &        53.1 &        54.8 &        61.1 &         65.6 &         68.4 &          72.3 &          72.1 &          76.1 &           73.4 &           74.4 &           74.9 \\
\cosmosqa{}    &       24.6 &        31.0 &        26.2 &        31.9 &         43.0 &         61.3 &          71.7 &          75.6 &          76.6 &           79.2 &           80.0 &           79.6 \\
\hellaswag{}   &       24.6 &        26.6 &        26.0 &        32.6 &         48.6 &         64.6 &          72.5 &          75.8 &          77.5 &           78.2 &           79.2 &           80.0 \\
\physicaliqa{} &       50.1 &        50.7 &        51.7 &        52.8 &         52.7 &         59.8 &          70.7 &          76.3 &          77.0 &           78.4 &           80.0 &           80.0 \\
\socialiqa{}   &       33.6 &        34.7 &        34.4 &        34.7 &         35.3 &         54.0 &          66.4 &          64.7 &          68.6 &           70.6 &           70.9 &           71.0 \\
\winogrande{}  &       50.1 &        50.7 &        52.9 &        50.7 &         51.1 &         57.9 &          64.2 &          67.4 &          69.9 &           71.4 &           72.2 &           73.1 \\
\bottomrule
\end{tabular}

    \caption{Learning curves for the single task baselines.}
    \label{tab:single-task_learning-curves}
\end{table*}

\begin{table*}
    \centering
    \begin{tabular}{llrrrrrrrrrrrr}
\toprule
 & &\multicolumn{12}{c}{\small{Size}}\\ \cline{3-14}
multiset & transfer &  \small{4} &  \small{10} &  \small{30} &  \small{91} &  \small{280} &  \small{865} &  \small{2667} &  \small{5334} &  \small{8000} &  \small{10667} &  \small{13334} &  \small{16000} \\
\midrule
\multirow{3}{*}{\glue{}} & fine-tune &       49.2 &        53.7 &        60.1 &        61.9 &         66.2 &         69.2 &          72.6 &          72.7 &          75.4 &           73.2 &           74.2 &           74.6 \\
             & multitask &       49.2 &        53.3 &        59.3 &        61.7 &         66.1 &         69.5 &          72.3 &          72.4 &          74.5 &           73.2 &           74.2 &           74.3 \\
             & sequential &       49.1 &        57.1 &        63.1 &        63.8 &         67.4 &         70.0 &          72.8 &          73.4 &          75.8 &           74.3 &           74.3 &           75.4 \\
\cline{1-14}
\multirow{3}{*}{\rainbow{}} & fine-tune &       61.0 &        63.4 &        69.6 &        71.3 &         72.6 &         73.9 &          76.6 &          75.8 &          76.7 &           75.8 &           76.4 &           76.7 \\
             & multitask &       61.9 &        62.9 &        69.6 &        71.4 &         73.0 &         74.3 &          76.4 &          76.7 &          76.9 &           76.8 &           77.5 &           77.2 \\
             & sequential &       61.4 &        65.3 &        70.9 &        71.0 &         73.6 &         73.8 &          75.8 &          75.7 &          76.6 &           76.6 &           76.6 &           76.5 \\
\cline{1-14}
\multirow{3}{*}{\superglue{}} & fine-tune &       49.1 &        57.0 &        59.9 &        63.6 &         66.1 &         69.1 &          71.3 &          71.9 &          75.3 &           73.6 &           73.3 &           74.5 \\
             & multitask &       49.1 &        57.0 &        59.5 &        63.7 &         65.5 &         67.9 &          71.3 &          71.6 &          74.3 &           72.8 &           72.7 &           74.5 \\
             & sequential &       49.2 &        61.5 &        62.6 &        64.8 &         66.3 &         70.2 &          72.2 &          72.3 &          76.1 &           73.6 &           74.0 &           75.2 \\
\bottomrule
\end{tabular}

    \caption{Learning curves on \anli{} using transfer.}
    \label{tab:anli_learning-curves}
\end{table*}

\begin{table*}
    \centering
    \begin{tabular}{llrrrrrrrrrrrr}
\toprule
 & &\multicolumn{12}{c}{\small{Size}}\\ \cline{3-14}
multiset & transfer &  \small{4} &  \small{10} &  \small{30} &  \small{91} &  \small{280} &  \small{865} &  \small{2667} &  \small{5334} &  \small{8000} &  \small{10667} &  \small{13334} &  \small{16000} \\
\midrule
\multirow{3}{*}{\glue{}} & fine-tune &       24.5 &        32.4 &        26.0 &        29.3 &         41.5 &         60.7 &          71.9 &          75.7 &          76.3 &           78.6 &           79.8 &           80.1 \\
             & multitask &       24.7 &        32.4 &        26.2 &        29.1 &         41.4 &         59.8 &          71.5 &          75.5 &          76.1 &           77.8 &           78.9 &           79.0 \\
             & sequential &       24.5 &        28.2 &        25.4 &        27.2 &         33.1 &         59.8 &          71.5 &          75.5 &          77.3 &           79.0 &           79.4 &           79.8 \\
\cline{1-14}
\multirow{3}{*}{\rainbow{}} & fine-tune &       54.9 &        61.9 &        57.4 &        60.9 &         62.1 &         67.4 &          74.7 &          78.2 &          79.1 &           80.1 &           81.2 &           81.3 \\
             & multitask &       53.3 &        61.8 &        57.7 &        61.4 &         62.5 &         66.3 &          74.6 &          76.9 &          77.8 &           77.3 &           80.0 &           80.6 \\
             & sequential &       58.2 &        60.1 &        57.7 &        61.6 &         65.0 &         68.7 &          76.4 &          78.1 &          78.7 &           80.2 &           80.1 &           80.5 \\
\cline{1-14}
\multirow{3}{*}{\superglue{}} & fine-tune &       24.6 &        35.3 &        26.0 &        41.5 &         46.4 &         61.0 &          70.6 &          75.6 &          76.7 &           78.8 &           79.9 &           80.5 \\
             & multitask &       24.6 &        34.4 &        26.0 &        40.6 &         44.8 &         60.3 &          70.9 &          74.4 &          75.4 &           77.8 &           77.9 &           78.8 \\
             & sequential &       26.4 &        38.0 &        33.6 &        50.1 &         49.3 &         60.4 &          71.3 &          75.2 &          75.6 &           78.3 &           80.0 &           80.0 \\
\bottomrule
\end{tabular}

    \caption{Learning curves on \cosmosqa{} using transfer.}
    \label{tab:cosmosqa_learning-curves}
\end{table*}

\begin{table*}
    \centering
    \begin{tabular}{llrrrrrrrrrrrr}
\toprule
 & &\multicolumn{12}{c}{\small{Size}}\\ \cline{3-14}
multiset & transfer &  \small{4} &  \small{10} &  \small{30} &  \small{91} &  \small{280} &  \small{865} &  \small{2667} &  \small{5334} &  \small{8000} &  \small{10667} &  \small{13334} &  \small{16000} \\
\midrule
\multirow{3}{*}{\glue{}} & fine-tune &       25.3 &        26.4 &        26.7 &        35.3 &         50.2 &         64.3 &          72.7 &          75.2 &          77.0 &           77.9 &           78.8 &           79.6 \\
             & multitask &       25.1 &        26.4 &        26.8 &        36.0 &         49.3 &         63.8 &          72.1 &          75.1 &          76.9 &           77.3 &           78.2 &           78.9 \\
             & sequential &       26.1 &        26.3 &        27.0 &        33.3 &         39.8 &         64.6 &          72.7 &          75.4 &          77.1 &           77.7 &           78.8 &           79.7 \\
\cline{1-14}
\multirow{3}{*}{\rainbow{}} & fine-tune &       53.7 &        49.1 &        47.0 &        55.4 &         63.3 &         67.0 &          74.0 &          76.4 &          77.9 &           78.3 &           79.7 &           80.2 \\
             & multitask &       53.8 &        50.0 &        46.5 &        54.3 &         62.4 &         66.2 &          73.0 &          76.0 &          77.1 &           77.8 &           78.8 &           79.7 \\
             & sequential &       51.6 &        50.8 &        54.4 &        60.1 &         65.8 &         69.2 &          74.7 &          76.3 &          78.3 &           78.4 &           79.8 &           80.2 \\
\cline{1-14}
\multirow{3}{*}{\superglue{}} & fine-tune &       24.2 &        26.4 &        25.8 &        36.1 &         49.6 &         64.7 &          72.4 &          75.2 &          77.1 &           77.8 &           78.7 &           79.6 \\
             & multitask &       25.3 &        26.5 &        25.9 &        35.1 &         48.0 &         63.3 &          71.5 &          74.4 &          76.5 &           77.0 &           77.4 &           78.9 \\
             & sequential &       25.8 &        26.9 &        27.0 &        42.1 &         54.9 &         63.8 &          72.2 &          75.3 &          76.9 &           77.7 &           78.8 &           79.7 \\
\bottomrule
\end{tabular}

    \caption{Learning curves on \hellaswag{} using transfer.}
    \label{tab:hellaswag_learning-curves}
\end{table*}

\begin{table*}
    \centering
    \begin{tabular}{llrrrrrrrrrrrr}
\toprule
 & &\multicolumn{12}{c}{\small{Size}}\\ \cline{3-14}
multiset & transfer &  \small{4} &  \small{10} &  \small{30} &  \small{91} &  \small{280} &  \small{865} &  \small{2667} &  \small{5334} &  \small{8000} &  \small{10667} &  \small{13334} &  \small{16000} \\
\midrule
\multirow{3}{*}{\glue{}} & fine-tune &       50.3 &        49.8 &        53.5 &        52.8 &         53.3 &         53.8 &          70.3 &          75.0 &          77.0 &           77.4 &           79.4 &           79.9 \\
             & multitask &       50.3 &        49.6 &        53.8 &        52.7 &         53.6 &         53.4 &          69.5 &          74.0 &          75.5 &           76.0 &           77.6 &           78.9 \\
             & sequential &       50.3 &        49.8 &        53.2 &        51.9 &         56.8 &         59.7 &          71.2 &          74.9 &          77.4 &           78.3 &           79.3 &           80.6 \\
\cline{1-14}
\multirow{3}{*}{\rainbow{}} & fine-tune &       49.0 &        49.7 &        48.5 &        50.5 &         69.1 &         73.2 &          76.5 &          79.0 &          79.4 &           80.3 &           81.8 &           81.8 \\
             & multitask &       49.9 &        49.9 &        51.0 &        50.4 &         68.8 &         73.6 &          76.2 &          78.3 &          78.2 &           79.7 &           80.6 &           80.4 \\
             & sequential &       49.0 &        45.3 &        41.6 &        73.1 &         74.3 &         74.8 &          78.2 &          78.3 &          80.0 &           80.7 &           81.7 &           82.5 \\
\cline{1-14}
\multirow{3}{*}{\superglue{}} & fine-tune &       50.0 &        49.6 &        53.3 &        52.3 &         51.0 &         54.1 &          69.0 &          73.9 &          76.6 &           77.9 &           79.9 &           80.1 \\
             & multitask &       50.1 &        49.6 &        53.6 &        52.4 &         51.2 &         54.5 &          67.2 &          71.2 &          73.5 &           71.8 &           75.8 &           75.4 \\
             & sequential &       50.0 &        49.7 &        53.0 &        52.2 &         51.6 &         50.8 &          69.9 &          75.5 &          77.1 &           78.6 &           78.9 &           80.9 \\
\bottomrule
\end{tabular}

    \caption{Learning curves on \physicaliqa{} using transfer.}
    \label{tab:physicaliqa_learning-curves}
\end{table*}

\begin{table*}
    \centering
    \begin{tabular}{llrrrrrrrrrrrr}
\toprule
 & &\multicolumn{12}{c}{\small{Size}}\\ \cline{3-14}
multiset & transfer &  \small{4} &  \small{10} &  \small{30} &  \small{91} &  \small{280} &  \small{865} &  \small{2667} &  \small{5334} &  \small{8000} &  \small{10667} &  \small{13334} &  \small{16000} \\
\midrule
\multirow{3}{*}{\glue{}} & fine-tune &       33.6 &        34.1 &        35.2 &        35.1 &         34.4 &         48.1 &          66.6 &          67.5 &          68.7 &           71.8 &           71.2 &           72.8 \\
             & multitask &       33.6 &        34.7 &        35.2 &        34.9 &         34.8 &         47.0 &          61.6 &          67.5 &          67.7 &           70.8 &           70.4 &           72.0 \\
             & sequential &       33.6 &        34.6 &        34.1 &        35.1 &         35.9 &         53.5 &          68.4 &          68.2 &          70.1 &           71.0 &           72.9 &           72.4 \\
\cline{1-14}
\multirow{3}{*}{\rainbow{}} & fine-tune &       33.6 &        48.2 &        58.1 &        63.9 &         64.7 &         66.6 &          70.2 &          70.6 &          73.1 &           72.7 &           73.6 &           73.6 \\
             & multitask &       33.6 &        50.7 &        58.1 &        64.3 &         65.3 &         67.0 &          69.7 &          70.6 &          72.0 &           72.5 &           73.5 &           73.4 \\
             & sequential &       33.6 &        65.0 &        64.2 &        65.2 &         67.1 &         67.8 &          70.1 &          70.6 &          71.5 &           72.8 &           74.1 &           73.9 \\
\cline{1-14}
\multirow{3}{*}{\superglue{}} & fine-tune &       33.6 &        34.1 &        34.1 &        35.2 &         34.9 &         58.1 &          67.7 &          67.6 &          70.2 &           71.6 &           72.3 &           72.4 \\
             & multitask &       33.6 &        34.7 &        34.6 &        35.2 &         35.4 &         57.3 &          66.3 &          66.7 &          69.7 &           70.5 &           70.0 &           70.9 \\
             & sequential &       33.6 &        34.1 &        34.6 &        34.6 &         35.9 &         58.9 &          67.0 &          68.7 &          69.4 &           71.8 &           72.2 &           72.6 \\
\bottomrule
\end{tabular}

    \caption{Learning curves on \socialiqa{} using transfer.}
    \label{tab:socialiqa_learning-curves}
\end{table*}

\begin{table*}
    \centering
    \begin{tabular}{llrrrrrrrrrrrr}
\toprule
 & &\multicolumn{12}{c}{\small{Size}}\\ \cline{3-14}
multiset & transfer &  \small{4} &  \small{10} &  \small{30} &  \small{91} &  \small{280} &  \small{865} &  \small{2667} &  \small{5334} &  \small{8000} &  \small{10667} &  \small{13334} &  \small{16000} \\
\midrule
\multirow{3}{*}{\glue{}} & fine-tune &       48.5 &        50.5 &        52.3 &        50.4 &         51.5 &         59.4 &          64.8 &          67.2 &          69.9 &           71.3 &           72.5 &           74.2 \\
             & multitask &       49.2 &        50.8 &        51.9 &        51.7 &         51.7 &         59.5 &          63.8 &          66.1 &          68.7 &           69.4 &           71.0 &           71.7 \\
             & sequential &       48.6 &        50.5 &        52.3 &        49.9 &         52.2 &         58.8 &          65.4 &          67.2 &          69.6 &           71.1 &           72.8 &           73.0 \\
\cline{1-14}
\multirow{3}{*}{\rainbow{}} & fine-tune &       52.6 &        52.6 &        52.6 &        53.0 &         56.5 &         63.2 &          66.5 &          69.2 &          71.3 &           72.5 &           73.8 &           73.9 \\
             & multitask &       51.8 &        52.7 &        52.9 &        53.5 &         55.8 &         62.6 &          64.9 &          67.9 &          69.4 &           70.1 &           70.6 &           70.3 \\
             & sequential &       51.0 &        53.2 &        54.1 &        54.7 &         58.8 &         63.3 &          66.9 &          68.4 &          70.5 &           72.5 &           73.4 &           74.9 \\
\cline{1-14}
\multirow{3}{*}{\superglue{}} & fine-tune &       49.3 &        50.7 &        52.2 &        50.7 &         52.9 &         61.7 &          65.2 &          67.2 &          70.1 &           72.1 &           73.5 &           73.5 \\
             & multitask &       50.7 &        50.7 &        52.1 &        51.6 &         52.2 &         60.1 &          64.3 &          64.9 &          68.0 &           68.5 &           70.0 &           69.8 \\
             & sequential &       48.8 &        50.4 &        52.4 &        51.7 &         52.9 &         60.6 &          64.5 &          66.6 &          69.0 &           71.3 &           72.1 &           73.2 \\
\bottomrule
\end{tabular}

    \caption{Learning curves on \winogrande{} using transfer.}
    \label{tab:winogrande_learning-curves}
\end{table*}

\subsection{Transferring to Other Tasks}
\label{app:experiments:transferring-to-other-tasks}

These experiments target \commonsenseqa{} and \joci{}.

\begin{figure*}[h]
    \centering
    \includegraphics[width=0.7\textwidth]{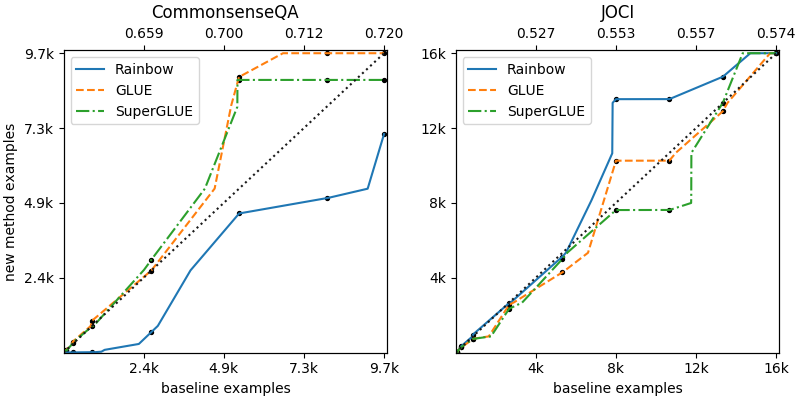}
    \caption{A comparison of transfer from different multisets to \commonsenseqa{} and \joci{} with \tfive{-Large} via multitask training. Performance is dev accuracy.}
    \label{fig:compare-multisets_external-tasks}
\end{figure*}

\begin{description}
\item{\textbf{Figure~\ref{fig:compare-multisets_external-tasks}}} A comparison of different multisets using multitask training
\item{\textbf{Table~\ref{tab:single-task_full-tasks_external}}} Single task baselines using the full training data
\item{\textbf{Table~\ref{tab:multiset_full-tasks_external}}} The performance using transfer and the full training data
\item{\textbf{Table~\ref{tab:single-task_learning-curves_external}}} Single task learning curves
\item{\textbf{Table~\ref{tab:multiset_learning-curves_external}}} Learning curves using transfer
\end{description}


\begin{table*}
    \centering
    \begin{tabular}{lrr}
\toprule
 &\multicolumn{2}{c}{\small{Task}}\\ \cline{2-3}
model &  \small{\commonsenseqa{}} &  \small{\joci{}} \\
\midrule
large &                      71.6 &             58.0 \\
\bottomrule
\end{tabular}

    \caption{Single task baselines using the full training data.}
    \label{tab:single-task_full-tasks_external}
\end{table*}

\begin{table*}
    \centering
    \begin{tabular}{lrr}
\toprule
 &\multicolumn{2}{c}{\small{Task}}\\ \cline{2-3}
multiset     &  \small{\commonsenseqa{}} &  \small{\joci{}} \\
\midrule
\glue{}      &                      70.8 &             57.8 \\
\rainbow{}   &                      72.6 &             57.5 \\
\superglue{} &                      70.5 &             58.3 \\
\bottomrule
\end{tabular}

    \caption{The performance on \commonsenseqa{} and \joci{} using transfer via multitask training.}
    \label{tab:multiset_full-tasks_external}
\end{table*}

\begin{table*}
    \centering
    \begin{tabular}{lrrrrrrrrrrrrr}
\toprule
 &\multicolumn{13}{c}{\small{Size}}\\ \cline{2-14}
task             &  \small{4} &  \small{10} &  \small{30} &  \small{91} &  \small{280} &  \small{865} &  \small{2667} &  \small{5334} &  \small{8000} &  \small{9741} &  \small{10667} &  \small{13334} &  \small{16000} \\
\midrule
\commonsenseqa{} &       19.9 &        35.1 &        45.6 &        53.6 &         58.3 &         63.2 &          66.3 &          70.8 &          71.4 &          72.0 &            -- &            -- &            -- \\
\joci{}          &       21.8 &        24.6 &        29.3 &        28.8 &         43.3 &         48.7 &          52.0 &          53.5 &          55.4 &           -- &           55.3 &           56.0 &           57.4 \\
\bottomrule
\end{tabular}

    \caption{Learning curves for the single task baselines on \commonsenseqa{} and \joci{}.}
    \label{tab:single-task_learning-curves_external}
\end{table*}

\begin{table*}
    \centering
    \begin{tabular}{p{1.25cm}lrrrrrrrrrrrrr}
\toprule
 & &\multicolumn{13}{c}{\small{Size}}\\ \cline{3-15}
task & multiset &  \small{4} &  \small{10} &  \small{30} &  \small{91} &  \small{280} &  \small{865} &  \small{2667} &  \small{5334} &  \small{8000} &  \small{9741} &  \small{10667} &  \small{13334} &  \small{16000} \\
\midrule
\multirow{3}{*}{\shortstack[l]{\small{\dataset{Common-}} \\ \small{\dataset{senseQA}}}} & \small{\glue{}} &       21.5 &        31.0 &        42.3 &        53.5 &         57.7 &         62.9 &          66.3 &          69.5 &          70.4 &          71.1 &            -- &            -- &            -- \\
        & \small{\rainbow{}} &       41.7 &        63.2 &        63.7 &        63.9 &         65.7 &         66.7 &          68.3 &          71.8 &          72.1 &          73.0 &         --    &             -- &            -- \\
        & \small{\superglue{}} &       20.7 &        35.0 &        42.0 &        54.1 &         57.9 &         63.3 &          65.9 &          69.0 &          70.7 &          70.7 &            -- &            -- &            -- \\
\cline{1-15}
\multirow{3}{*}{\small{\joci{}}} & \small{\glue{}} &       22.4 &        24.7 &        30.5 &        29.0 &         43.4 &         50.2 &          52.1 &          54.4 &          54.9 &           -- &           55.4 &           56.1 &           57.2 \\
        & \small{\rainbow{}} &       21.8 &        24.2 &        30.2 &        30.3 &         42.6 &         48.5 &          52.0 &          53.6 &          54.4 &           -- &           55.4 &           55.0 &           56.7 \\
        & \small{\superglue{}} &       21.9 &        24.5 &        30.2 &        29.2 &         43.1 &         50.4 &          52.4 &          53.6 &          55.8 &           -- &           55.4 &           56.0 &           56.5 \\
\bottomrule
\end{tabular}

    \caption{Learning curves using multitask training on \commonsenseqa{} and \joci{}.}
    \label{tab:multiset_learning-curves_external}
\end{table*}

\subsection{Effect of Size}
\label{app:experiments:effect-of-size}

These experiments explore the impact of model size on transfer using \commonsenseqa{} as the target dataset.

\begin{figure*}[h]
    \centering
    \includegraphics[width=0.85\textwidth]{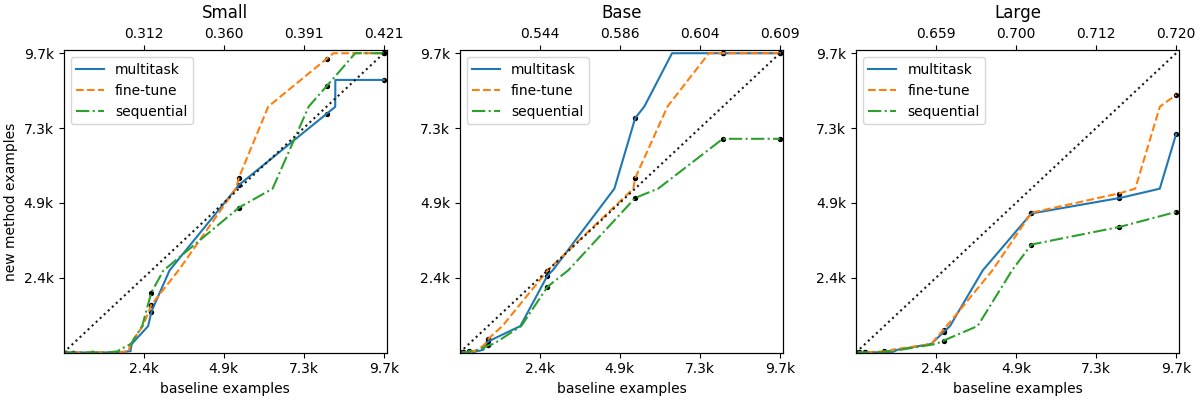}
    \caption{A comparison of transfer methods from \rainbow{} to \commonsenseqa{} across model sizes with \tfive{-Small}, \tfive{-Base}, and \tfive{-Large}. Performance is dev accuracy.}
    \label{fig:effect-of-size}
\end{figure*}

\begin{description}
\item{\textbf{Figure~\ref{fig:effect-of-size}}} A comparison of transfer methods across different model sizes
\item{\textbf{Table~\ref{tab:effect-of-size_mixtures}}} Full task performance for the initial multitask models used in sequential training and multitask fine-tune training experiments comparing model size
\item{\textbf{Table~\ref{tab:single-task_learning-curves_size}}} Single task learning curves across different model sizes
\item{\textbf{Table~\ref{tab:multiset_learning-curves_size}}} Learning curves using transfer across different model sizes
\end{description}

\begin{table*}
    \centering
    \begin{tabular}{lrrrrrr}
\toprule
 &\multicolumn{6}{c}{\small{Task}}\\ \cline{2-7}
model &  \small{\anli{}} &  \small{\cosmosqa{}} &  \small{\hellaswag{}} &  \small{\physicaliqa{}} &  \small{\socialiqa{}} &  \small{\winogrande{}} \\
\midrule
base  &             65.3 &                 72.8 &                  56.2 &                    73.3 &                  66.1 &                   61.8 \\
large &             76.2 &                 81.1 &                  81.3 &                    80.7 &                  74.5 &                   72.1 \\
small &             57.0 &                 44.5 &                  31.8 &                    54.6 &                  46.8 &                   52.4 \\
\bottomrule
\end{tabular}

    \caption{Full task performance for \universalmodel{} on \rainbow{} after multitask training and before training on the target dataset (\commonsenseqa{}) across different model sizes.}
    \label{tab:effect-of-size_mixtures}
\end{table*}

\begin{table*}
    \centering
    \begin{tabular}{lrrrrrrrrrr}
\toprule
 &\multicolumn{10}{c}{\small{Size}}\\ \cline{2-11}
model &  \small{4} &  \small{10} &  \small{30} &  \small{91} &  \small{280} &  \small{865} &  \small{2667} &  \small{5334} &  \small{8000} &  \small{9741} \\
\midrule
base  &       29.1 &        30.6 &        29.2 &        41.5 &         46.4 &         49.7 &          55.1 &          59.3 &          60.9 &          60.8 \\
large &       19.9 &        35.1 &        45.6 &        53.6 &         58.3 &         63.2 &          66.3 &          70.8 &          71.4 &          72.0 \\
small &       20.5 &        23.1 &        19.2 &        26.4 &         25.4 &         24.9 &          32.0 &          36.8 &          40.0 &          42.1 \\
\bottomrule
\end{tabular}

    \caption{Learning curves for single task baselines on \commonsenseqa{} at different model sizes.}
    \label{tab:single-task_learning-curves_size}
\end{table*}

\begin{table*}
    \centering
    \begin{tabular}{llrrrrrrrrrr}
\toprule
 & &\multicolumn{10}{c}{\small{Size}}\\ \cline{3-12}
model & transfer &  \small{4} &  \small{10} &  \small{30} &  \small{91} &  \small{280} &  \small{865} &  \small{2667} &  \small{5334} &  \small{8000} &  \small{9741} \\
\midrule
\multirow{3}{*}{base} & fine-tune &       37.5 &        47.1 &        46.5 &        47.9 &         49.2 &         51.0 &          55.1 &          59.2 &          59.9 &          60.6 \\
      & multitask &       36.4 &        47.7 &        46.9 &        48.9 &         49.2 &         52.7 &          55.4 &          58.3 &          59.5 &          60.0 \\
      & sequential &       38.4 &        45.4 &        45.6 &        48.1 &         50.2 &         52.7 &          56.1 &          59.7 &          61.6 &          61.6 \\
\cline{1-12}
\multirow{3}{*}{large} & fine-tune &       47.4 &        64.0 &        63.8 &        63.0 &         65.7 &         66.5 &          68.8 &          71.6 &          71.8 &          72.6 \\
      & multitask &       41.7 &        63.2 &        63.7 &        63.9 &         65.7 &         66.7 &          68.3 &          71.8 &          72.1 &          73.0 \\
      & sequential &       59.5 &        63.3 &        63.0 &        64.0 &         65.9 &         68.1 &          69.8 &          72.9 &          73.1 &          72.6 \\
\cline{1-12}
\multirow{3}{*}{small} & fine-tune &       26.3 &        30.9 &        30.3 &        27.7 &         29.7 &         31.1 &          33.5 &          36.6 &          37.8 &          40.2 \\
      & multitask &       26.3 &        30.7 &        29.7 &        28.9 &         29.9 &         31.7 &          33.0 &          36.6 &          40.5 &          40.1 \\
      & sequential &       25.1 &        28.7 &        29.1 &        27.8 &         29.8 &         31.0 &          32.7 &          38.0 &          39.3 &          41.0 \\
\bottomrule
\end{tabular}

    \caption{Learning curves for \universalmodel{} on \commonsenseqa{} at different model sizes, with different transfer approaches.}
    \label{tab:multiset_learning-curves_size}
\end{table*}

\subsection{Transferring Knowledge Graphs}
\label{app:experiments:transferring-knowledge-graphs}

These experiments explore transferring knowledge graphs via multitask training, using \rainbow{} for the end tasks.

\begin{figure*}[h]
    \centering
    \includegraphics[width=0.85\textwidth]{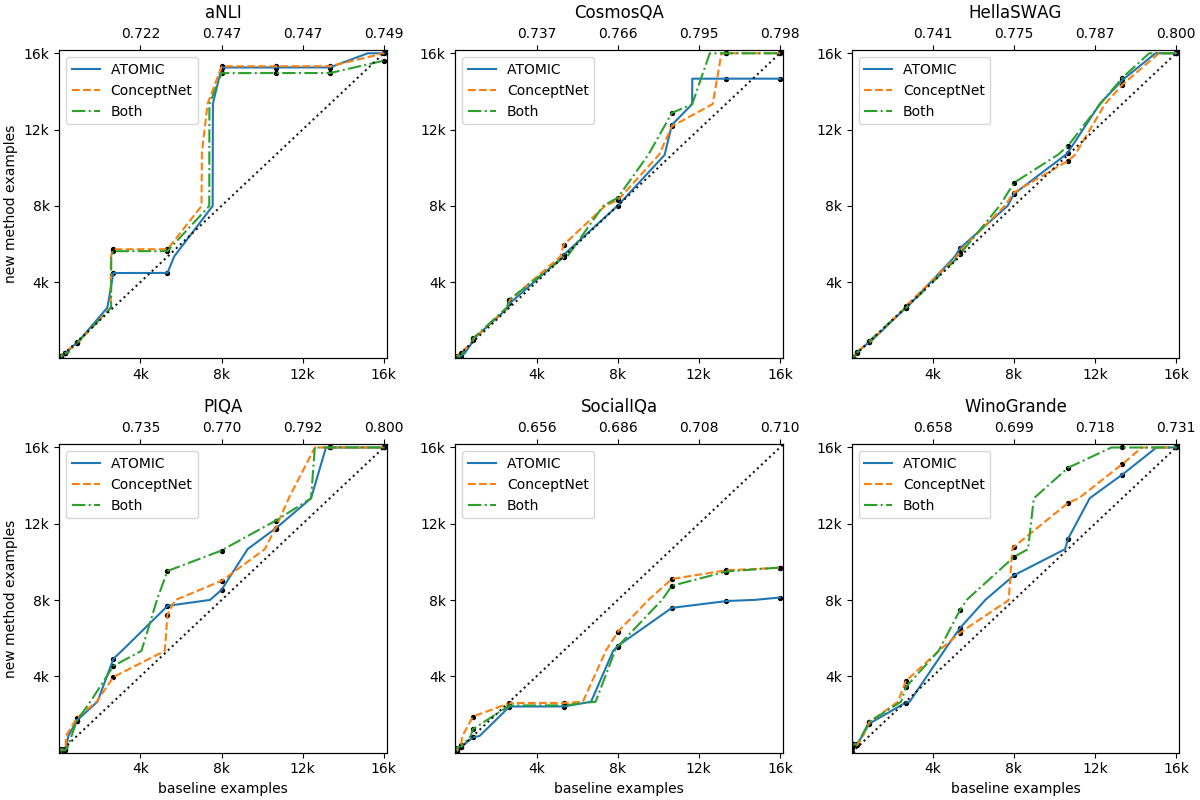}
    \caption{A comparison of transfer from different knowledge graphs to the \rainbow{} tasks using multitask training. Performance is dev accuracy.}
    \label{fig:comparing-knowledge-graphs_no-rainbow}
\end{figure*}

\begin{figure*}[h]
    \centering
    \includegraphics[width=0.85\textwidth]{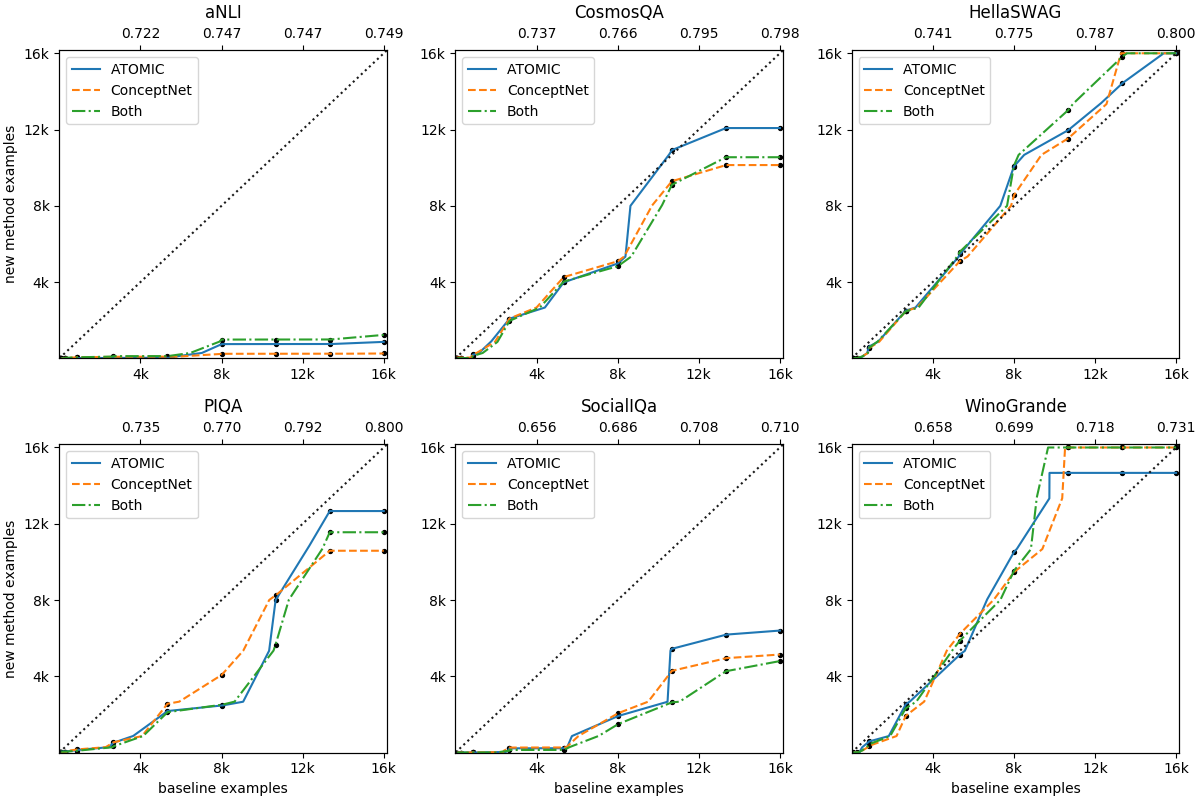}
    \caption{A comparison of transfer from different knowledge graphs and \rainbow{} to the \rainbow{} tasks using multitask training. Performance is dev accuracy.}
    \label{fig:comparing-knowledge-graphs_rainbow}
\end{figure*}

\begin{figure*}[h]
    \centering
    \includegraphics[width=0.85\textwidth]{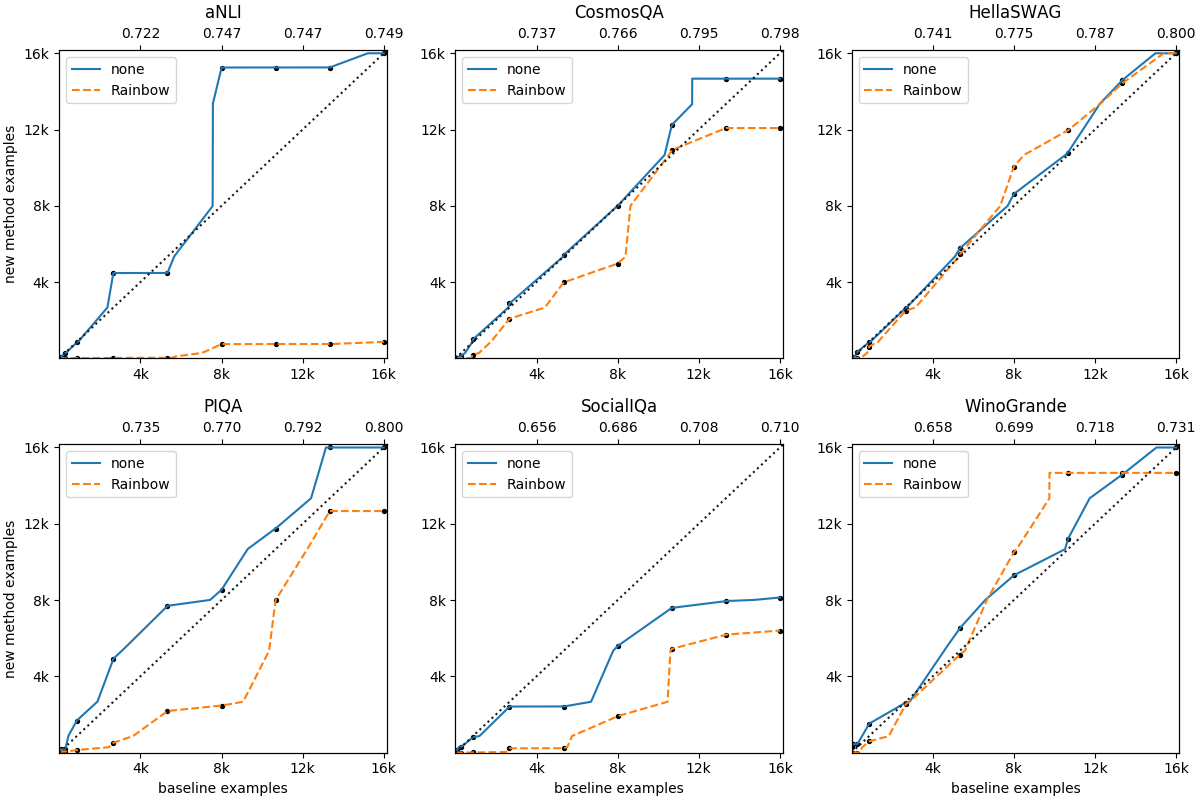}
    \caption{A comparison of transfer from \atomic{} to the \rainbow{} tasks via multitask training when also and not also multitasking with \rainbow{}.}
    \label{fig:comparing-knowledge-graphs-w-multisets_atomic}
\end{figure*}

\begin{figure*}[h]
    \centering
    \includegraphics[width=0.85\textwidth]{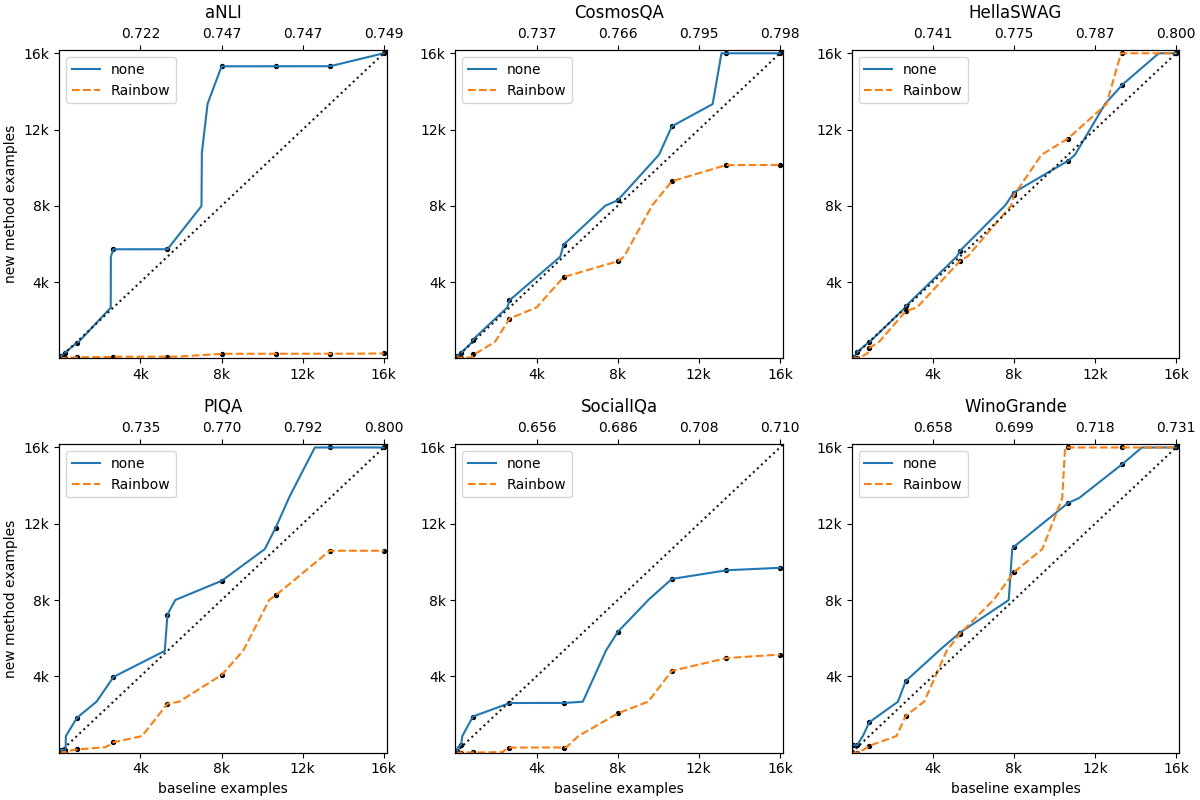}
    \caption{A comparison of transfer from \conceptnet{} to the \rainbow{} tasks via multitask training when also and not also multitasking with \rainbow{}.}
    \label{fig:comparing-knowledge-graphs-w-multisets_conceptnet}
\end{figure*}

\begin{figure*}[h]
    \centering
    \includegraphics[width=0.85\textwidth]{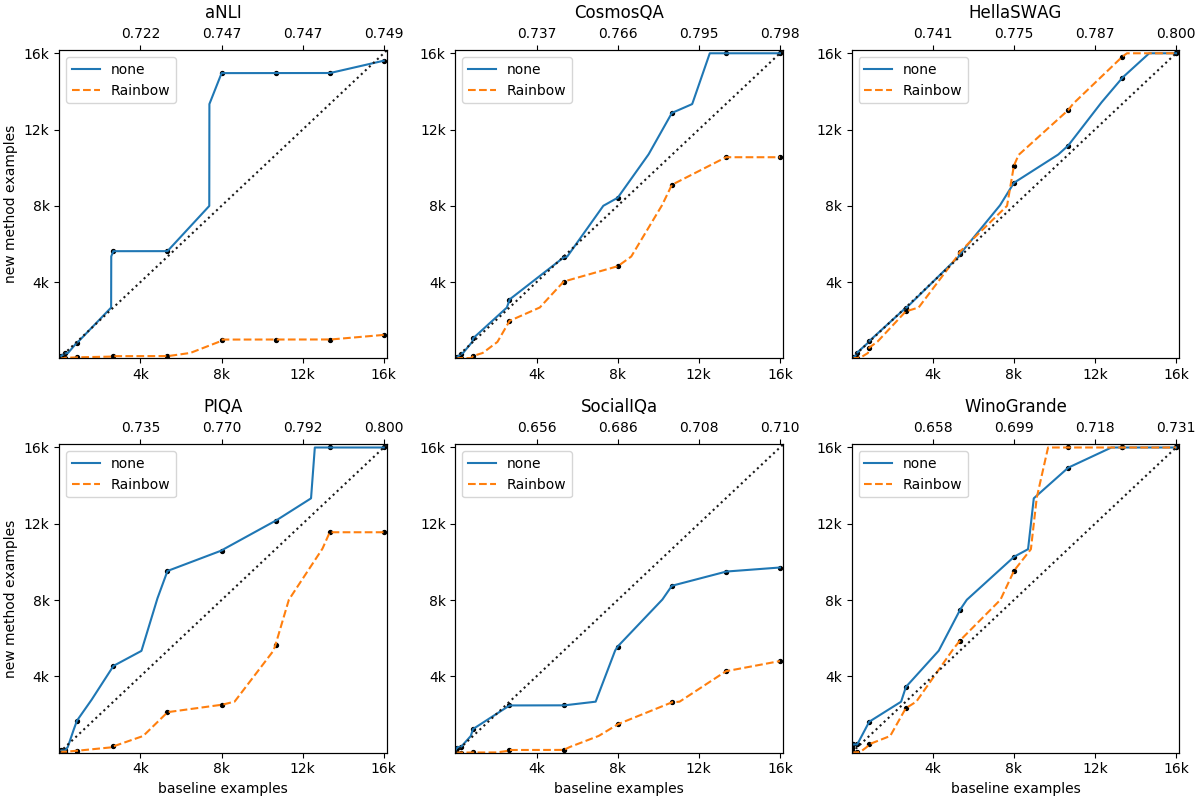}
    \caption{A comparison of transfer from both \atomic{} and \conceptnet{} to the \rainbow{} tasks via multitask training when also and not also multitasking with \rainbow{}.}
    \label{fig:comparing-knowledge-graphs-w-multisets_both}
\end{figure*}

\begin{description}
\item{\textbf{Figure~\ref{fig:comparing-knowledge-graphs_no-rainbow}}} A comparison of transfer from different knowledge graphs
\item{\textbf{Figure~\ref{fig:comparing-knowledge-graphs_rainbow}}} A comparison of transfer from different knowledge graphs when also multitasking with \rainbow{}
\item{\textbf{Figure~\ref{fig:comparing-knowledge-graphs-w-multisets_atomic}}} A comparison of transfer from \atomic{} with and without multitasking \rainbow{}
\item{\textbf{Figure~\ref{fig:comparing-knowledge-graphs-w-multisets_conceptnet}}} A comparison of transfer from \conceptnet{} with and without multitasking \rainbow{}
\item{\textbf{Figure~\ref{fig:comparing-knowledge-graphs-w-multisets_both}}} A comparison of transfer from both \atomic{} and \conceptnet{} with and without multitasking \rainbow{}
\item{\textbf{Table~\ref{tab:multiset_full-tasks_knowledge-graphs}}} The performance using transfer and the full training data
\item{\textbf{Table~\ref{tab:anli_learning-curves_knowledge-graphs}}} \anli{} learning curves with transfer
\item{\textbf{Table~\ref{tab:cosmosqa_learning-curves_knowledge-graphs}}} \cosmosqa{} learning curves with transfer
\item{\textbf{Table~\ref{tab:hellaswag_learning-curves_knowledge-graphs}}} \hellaswag{} learning curves with transfer
\item{\textbf{Table~\ref{tab:physicaliqa_learning-curves_knowledge-graphs}}} \physicaliqa{} learning curves with transfer
\item{\textbf{Table~\ref{tab:socialiqa_learning-curves_knowledge-graphs}}} \socialiqa{} learning curves with transfer
\item{\textbf{Table~\ref{tab:winogrande_learning-curves_knowledge-graphs}}} \winogrande{} learning curves with transfer
\end{description}


\begin{table*}
    \centering
    \begin{tabular}{lllrrrrrr}
\toprule
 & & &\multicolumn{6}{c}{\small{Task}}\\ \cline{4-9}
multiset & knowledge & direction &  \small{\anli{}} &  \small{\cosmosqa{}} &  \small{\hellaswag{}} &  \small{\physicaliqa{}} &  \small{\socialiqa{}} &  \small{\winogrande{}} \\
\midrule
\multirow{9}{*}{\none{}} & \multirow{3}{*}{\atomic{}} & backward &             78.3 &                 81.8 &                  82.5 &                    79.5 &                  74.3 &                   76.9 \\
           &             & bidirectional &             77.9 &                 81.0 &                  82.8 &                    79.3 &                  75.0 &                   78.2 \\
           &             & forward &             77.7 &                 81.0 &                  82.4 &                    79.9 &                  74.4 &                   76.8 \\
\cline{2-9}
           & \multirow{3}{*}{\conceptnet{}} & backward &             78.0 &                 81.8 &                  82.5 &                    79.4 &                  74.3 &                   76.3 \\
           &             & bidirectional &             77.8 &                 81.5 &                  82.3 &                    80.5 &                  74.1 &                   76.3 \\
           &             & forward &             77.5 &                 81.6 &                  82.1 &                    80.0 &                  73.7 &                   76.2 \\
\cline{2-9}
           & \multirow{3}{*}{\together{}} & backward &             78.0 &                 81.2 &                  82.4 &                    81.1 &                  74.4 &                   75.7 \\
           &             & bidirectional &             77.6 &                 81.0 &                  82.6 &                    80.1 &                  74.7 &                   76.4 \\
           &             & forward &             77.5 &                 81.8 &                  82.7 &                    79.8 &                  74.8 &                   76.6 \\
\cline{1-9}
\cline{2-9}
\multirow{9}{*}{\rainbow{}} & \multirow{3}{*}{\atomic{}} & backward &             78.3 &                 81.4 &                  81.3 &                    80.5 &                  75.0 &                   73.2 \\
           &             & bidirectional &             77.7 &                 81.4 &                  81.2 &                    80.4 &                  74.9 &                   71.5 \\
           &             & forward &             77.9 &                 81.3 &                  81.6 &                    80.4 &                  75.0 &                   73.6 \\
\cline{2-9}
           & \multirow{3}{*}{\conceptnet{}} & backward &             78.7 &                 81.8 &                  81.6 &                    81.3 &                  75.0 &                   73.5 \\
           &             & bidirectional &             78.8 &                 80.6 &                  81.5 &                    81.0 &                  74.7 &                   72.9 \\
           &             & forward &             78.3 &                 81.6 &                  81.3 &                    80.6 &                  75.5 &                   73.4 \\
\cline{2-9}
           & \multirow{3}{*}{\together{}} & backward &             77.1 &                 80.9 &                  81.6 &                    81.2 &                  74.3 &                   71.6 \\
           &             & bidirectional &             76.9 &                 81.4 &                  81.0 &                    80.3 &                  75.0 &                   72.2 \\
           &             & forward &             77.7 &                 81.9 &                  81.7 &                    80.8 &                  74.8 &                   72.1 \\
\bottomrule
\end{tabular}

    \caption{The performance when transferring different knowledge graphs to \rainbow{} with multitask training using the full training data.}
    \label{tab:multiset_full-tasks_knowledge-graphs}
\end{table*}

\begin{table*}
    \centering
    \begin{tabular}{llrrrrrrrrrrrr}
\toprule
 & &\multicolumn{12}{c}{\small{Size}}\\ \cline{3-14}
multiset & knowledge &  \small{4} &  \small{10} &  \small{30} &  \small{91} &  \small{280} &  \small{865} &  \small{2667} &  \small{5334} &  \small{8000} &  \small{10667} &  \small{13334} &  \small{16000} \\
\midrule
\multirow{3}{*}{\none{}} & \atomic{} &       49.3 &        52.3 &        54.6 &        61.6 &         65.8 &         68.5 &          71.6 &          72.5 &          75.3 &           73.4 &           74.1 &           74.8 \\
           & \conceptnet{} &       49.2 &        54.6 &        53.5 &        60.2 &         65.5 &         68.6 &          72.2 &          71.7 &          74.4 &           73.1 &           74.0 &           74.9 \\
           & \together{} &       49.3 &        54.0 &        52.9 &        59.6 &         66.3 &         68.5 &          72.4 &          71.6 &          74.7 &           73.6 &           74.0 &           75.0 \\
\cline{1-14}
\multirow{3}{*}{\rainbow{}} & \atomic{} &       55.0 &        67.0 &        73.0 &        72.0 &         73.8 &         74.9 &          76.2 &          76.4 &          76.6 &           76.8 &           76.6 &           77.0 \\
           & \conceptnet{} &       56.7 &        66.4 &        64.5 &        72.7 &         75.2 &         75.1 &          76.2 &          76.4 &          77.1 &           76.3 &           76.4 &           76.6 \\
           & \together{} &       58.3 &        64.7 &        65.4 &        72.1 &         73.2 &         74.5 &          76.2 &          76.0 &          76.6 &           76.6 &           77.0 &           76.7 \\
\bottomrule
\end{tabular}

    \caption{Learning curves on \anli{} using transfer from knowledge graphs via multitask training.}
    \label{tab:anli_learning-curves_knowledge-graphs}
\end{table*}

\begin{table*}
    \centering
    \begin{tabular}{llrrrrrrrrrrrr}
\toprule
 & &\multicolumn{12}{c}{\small{Size}}\\ \cline{3-14}
multiset & knowledge &  \small{4} &  \small{10} &  \small{30} &  \small{91} &  \small{280} &  \small{865} &  \small{2667} &  \small{5334} &  \small{8000} &  \small{10667} &  \small{13334} &  \small{16000} \\
\midrule
\multirow{3}{*}{\none{}} & \atomic{} &       24.6 &        34.8 &        25.5 &        40.9 &         49.1 &         60.6 &          71.4 &          75.6 &          76.6 &           78.8 &           79.4 &           79.4 \\
           & \conceptnet{} &       24.6 &        31.5 &        25.7 &        28.0 &         42.7 &         60.6 &          71.1 &          75.4 &          76.4 &           78.6 &           79.6 &           79.7 \\
           & \together{} &       24.7 &        29.0 &        24.9 &        33.2 &         46.0 &         60.3 &          71.0 &          75.7 &          76.3 &           78.1 &           79.4 &           79.6 \\
\cline{1-14}
\multirow{3}{*}{\rainbow{}} & \atomic{} &       59.8 &        59.2 &        56.5 &        59.5 &         63.0 &         66.4 &          74.3 &          77.0 &          77.2 &           79.0 &           80.7 &           80.2 \\
           & \conceptnet{} &       61.0 &        61.1 &        57.5 &        58.2 &         62.3 &         67.6 &          73.7 &          76.9 &          78.2 &           80.2 &           80.7 &           79.8 \\
           & \together{} &       59.7 &        60.3 &        57.7 &        60.7 &         64.1 &         68.3 &          73.9 &          77.3 &          78.7 &           79.8 &           80.6 &           79.1 \\
\bottomrule
\end{tabular}

    \caption{Learning curves on \cosmosqa{} using transfer from knowledge graphs via multitask training.}
    \label{tab:cosmosqa_learning-curves_knowledge-graphs}
\end{table*}

\begin{table*}
    \centering
    \begin{tabular}{llrrrrrrrrrrrr}
\toprule
 & &\multicolumn{12}{c}{\small{Size}}\\ \cline{3-14}
multiset & knowledge &  \small{4} &  \small{10} &  \small{30} &  \small{91} &  \small{280} &  \small{865} &  \small{2667} &  \small{5334} &  \small{8000} &  \small{10667} &  \small{13334} &  \small{16000} \\
\midrule
\multirow{3}{*}{\none{}} & \atomic{} &       25.5 &        26.4 &        26.0 &        37.4 &         46.8 &         64.8 &          72.5 &          75.5 &          77.3 &           78.2 &           78.8 &           79.7 \\
           & \conceptnet{} &       25.2 &        26.1 &        25.7 &        35.7 &         46.6 &         64.7 &          72.4 &          75.6 &          77.2 &           78.3 &           78.9 &           79.7 \\
           & \together{} &       25.9 &        26.1 &        26.5 &        37.4 &         47.8 &         64.4 &          72.5 &          75.7 &          77.1 &           78.1 &           78.8 &           79.6 \\
\cline{1-14}
\multirow{3}{*}{\rainbow{}} & \atomic{} &       54.1 &        45.9 &        49.3 &        55.5 &         61.9 &         66.5 &          73.1 &          75.7 &          77.1 &           77.7 &           78.8 &           79.8 \\
           & \conceptnet{} &       53.9 &        47.2 &        47.4 &        54.6 &         62.8 &         66.7 &          73.2 &          76.0 &          77.4 &           77.9 &           78.9 &           79.2 \\
           & \together{} &       54.4 &        45.0 &        49.1 &        55.0 &         63.0 &         66.3 &          73.3 &          75.6 &          77.3 &           77.6 &           78.3 &           79.3 \\
\bottomrule
\end{tabular}

    \caption{Learning curves on \hellaswag{} using transfer from knowledge graphs via multitask training.}
    \label{tab:hellaswag_learning-curves_knowledge-graphs}
\end{table*}

\begin{table*}
    \centering
    \begin{tabular}{llrrrrrrrrrrrr}
\toprule
 & &\multicolumn{12}{c}{\small{Size}}\\ \cline{3-14}
multiset & knowledge &  \small{4} &  \small{10} &  \small{30} &  \small{91} &  \small{280} &  \small{865} &  \small{2667} &  \small{5334} &  \small{8000} &  \small{10667} &  \small{13334} &  \small{16000} \\
\midrule
\multirow{3}{*}{\none{}} & \atomic{} &       51.0 &        50.4 &        54.0 &        50.8 &         53.2 &         54.7 &          66.0 &          71.6 &          76.9 &           77.7 &           79.4 &           79.9 \\
           & \conceptnet{} &       50.4 &        50.2 &        54.0 &        51.0 &         54.5 &         52.0 &          65.7 &          76.0 &          76.4 &           78.1 &           78.8 &           79.5 \\
           & \together{} &       50.5 &        50.1 &        53.9 &        51.1 &         54.4 &         56.6 &          63.9 &          73.6 &          75.2 &           77.1 &           79.4 &           79.5 \\
\cline{1-14}
\multirow{3}{*}{\rainbow{}} & \atomic{} &       50.0 &        48.9 &        50.5 &        56.7 &         69.4 &         72.7 &          77.6 &          78.2 &          78.4 &           79.3 &           80.2 &           80.6 \\
           & \conceptnet{} &       49.8 &        49.0 &        53.6 &        54.2 &         68.4 &         73.7 &          76.4 &          77.6 &          78.2 &           80.0 &           80.8 &           80.4 \\
           & \together{} &       50.1 &        49.1 &        50.0 &        59.5 &         70.5 &         73.8 &          77.4 &          78.3 &          78.8 &           79.8 &           80.4 &           80.4 \\
\bottomrule
\end{tabular}

    \caption{Learning curves on \physicaliqa{} using transfer from knowledge graphs via multitask training.}
    \label{tab:physicaliqa_learning-curves_knowledge-graphs}
\end{table*}

\begin{table*}
    \centering
    \begin{tabular}{llrrrrrrrrrrrr}
\toprule
 & &\multicolumn{12}{c}{\small{Size}}\\ \cline{3-14}
multiset & knowledge &  \small{4} &  \small{10} &  \small{30} &  \small{91} &  \small{280} &  \small{865} &  \small{2667} &  \small{5334} &  \small{8000} &  \small{10667} &  \small{13334} &  \small{16000} \\
\midrule
\multirow{3}{*}{\none{}} & \atomic{} &       33.6 &        34.2 &        33.5 &        35.0 &         34.5 &         56.0 &          67.1 &          68.3 &          71.0 &           72.0 &           72.6 &           73.2 \\
           & \conceptnet{} &       33.6 &        34.6 &        34.0 &        34.7 &         34.7 &         37.2 &          66.6 &          67.9 &          69.7 &           71.8 &           72.4 &           71.7 \\
           & \together{} &       33.6 &        34.4 &        33.7 &        35.1 &         34.1 &         50.6 &          67.3 &          68.4 &          70.2 &           71.5 &           72.1 &           72.3 \\
\cline{1-14}
\multirow{3}{*}{\rainbow{}} & \atomic{} &       33.6 &        54.8 &        64.3 &        65.0 &         65.8 &         66.0 &          70.4 &          70.5 &          71.8 &           72.2 &           73.2 &           73.8 \\
           & \conceptnet{} &       33.6 &        55.0 &        64.3 &        63.1 &         65.7 &         66.4 &          69.7 &          71.1 &          72.3 &           72.5 &           72.6 &           73.5 \\
           & \together{} &       33.6 &        61.4 &        62.3 &        65.5 &         65.9 &         67.5 &          70.6 &          71.1 &          71.8 &           72.5 &           73.5 &           74.4 \\
\bottomrule
\end{tabular}

    \caption{Learning curves on \socialiqa{} using transfer from knowledge graphs via multitask training.}
    \label{tab:socialiqa_learning-curves_knowledge-graphs}
\end{table*}

\begin{table*}
    \centering
    \begin{tabular}{llrrrrrrrrrrrr}
\toprule
 & &\multicolumn{12}{c}{\small{Size}}\\ \cline{3-14}
multiset & knowledge &  \small{4} &  \small{10} &  \small{30} &  \small{91} &  \small{280} &  \small{865} &  \small{2667} &  \small{5334} &  \small{8000} &  \small{10667} &  \small{13334} &  \small{16000} \\
\midrule
\multirow{3}{*}{\none{}} & \atomic{} &       50.4 &        50.5 &        52.3 &        49.6 &         49.6 &         54.1 &          64.4 &          66.5 &          68.6 &           71.3 &           71.7 &           72.8 \\
           & \conceptnet{} &       50.1 &        50.4 &        52.4 &        49.9 &         50.2 &         54.5 &          62.8 &          66.1 &          69.7 &           69.9 &           71.6 &           72.5 \\
           & \together{} &       50.7 &        50.3 &        51.9 &        50.4 &         49.9 &         53.9 &          63.4 &          66.1 &          67.7 &           70.3 &           70.5 &           72.1 \\
\cline{1-14}
\multirow{3}{*}{\rainbow{}} & \atomic{} &       51.9 &        53.5 &        52.8 &        52.9 &         54.1 &         61.2 &          64.4 &          67.6 &          68.7 &           70.0 &           71.0 &           70.8 \\
           & \conceptnet{} &       51.6 &        52.6 &        53.7 &        54.0 &         57.1 &         62.6 &          65.3 &          66.6 &          69.0 &           70.7 &           71.3 &           71.3 \\
           & \together{} &       50.4 &        52.6 &        52.0 &        53.9 &         56.5 &         61.5 &          64.8 &          66.9 &          69.3 &           70.4 &           70.6 &           70.9 \\
\bottomrule
\end{tabular}

    \caption{Learning curves on \winogrande{} using transfer from knowledge graphs via multitask training.}
    \label{tab:winogrande_learning-curves_knowledge-graphs}
\end{table*}

\end{document}